\documentclass[10pt,twocolumn,letterpaper]{article}
\pdfoutput=1
\usepackage{cvpr}
\usepackage{times}
\usepackage{enumitem}
\usepackage{pgffor}
\usepackage{graphicx}
\usepackage{amsmath}
\usepackage{amssymb}
\usepackage{overpic}
\usepackage{xcolor}
\usepackage{xspace}
\usepackage{multirow}
\usepackage[labelformat=empty]{subfig}
\usepackage[numbers,square,sort&compress]{natbib}
\usepackage{authblk}
\usepackage[
  pagebackref=true,
  breaklinks=true,
  letterpaper=true,
  colorlinks,
  bookmarks=false]{hyperref}
\usepackage{flushend}

\cvprfinalcopy
\ifcvprfinal\fi

\newcommand{\real}{\mathbb{R}}
\newcommand{\bx}{\mathbf{x}}
\newcommand{\bw}{\mathbf{w}}
\newcommand{\bu}{\mathbf{u}}
\newcommand{\bv}{\mathbf{v}}
\newcommand{\bd}{\mathbf{d}}

\newcommand{\DTDLIN}{$\text{DTD}_\text{LIN}$\xspace}
\newcommand{\DTDRBF}{$\text{DTD}_\text{RBF}$\xspace}
\newcommand{\IFVSIFT}{$\text{IFV}_\text{SIFT}$\xspace}
\newcommand{\IFVRGB}{$\text{IFV}_\text{RGB}$\xspace}

\begin{document}
\title{Describing Textures in the Wild}

\author[1]{Mircea Cimpoi}
\author[2]{Subhransu Maji}
\author[3]{Iasonas Kokkinos}
\author[4]{Sammy Mohamed}
\author[1]{Andrea Vedaldi}

\affil[1]{Department of Engineering Science, University of Oxford}
\affil[2]{Toyota Technological Institute, Chicago (TTIC)}
\affil[3]{Center for Visual Computing, Ecole Centrale Paris}
\affil[4]{Stony Brook University}

\maketitle
\thispagestyle{empty}

\begin{abstract}
Patterns and textures are defining characteristics of many natural objects: a shirt can be striped, the wings of a butterfly can be veined, and the skin of an animal can be scaly. Aiming at supporting this analytical dimension in image understanding, we address the challenging problem of \emph{describing textures} with semantic attributes. We identify a rich vocabulary of forty-seven texture terms and use them to describe a large dataset of patterns collected ``in the wild''. The resulting \emph{Describable Textures Dataset} (DTD) is the basis to seek for the best texture representation for recognizing describable texture attributes in images. We port from object recognition to texture recognition the Improved Fisher Vector (IFV) and show that, surprisingly, it outperforms specialized texture descriptors not only on our problem, but also in established material recognition datasets. We also show that the describable attributes are excellent  texture descriptors, transferring between datasets and tasks; in particular, combined with IFV, they significantly outperform the state-of-the-art by more than 8\% on both FMD  and KTH-TIPS-2b benchmarks. We also demonstrate that they produce intuitive descriptions of materials and Internet images.
\end{abstract}

\section{Introduction}

\begin{figure}[!t]
\centering
\includegraphics[width=0.99\linewidth]{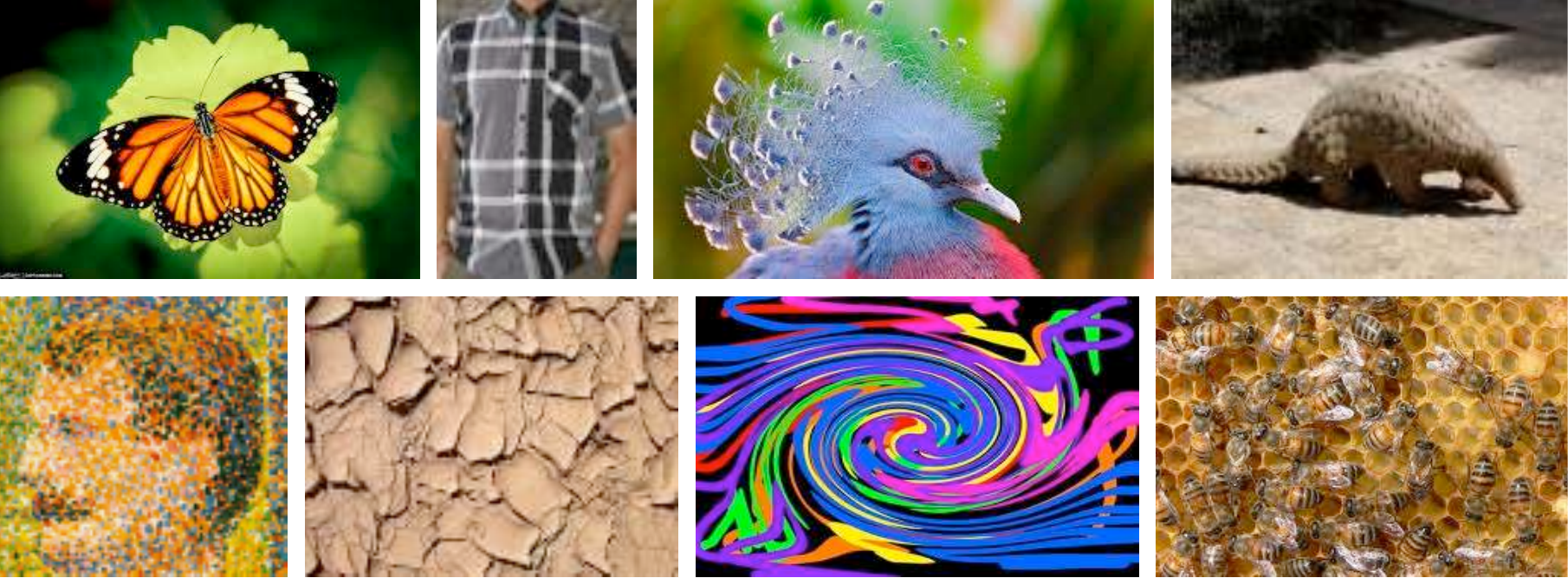}
\caption{\label{fig:examples} Both the man-made and the natural world are an abundant source of richly textured objects. The textures of objects shown above can be described (in no particular order) as dotted, striped, chequered, cracked, swirly, honeycombed, and scaly.
We aim at identifying these attributes automatically and generating descriptions based on them.}
\end{figure}

Recently \emph{visual attributes} have raised significant interest in the community~\cite{farhadi2009describing,patterson2012sun,bourdev2011describing,kumar2011describable}. A ``visual attribute'' is a property of an object that can be measured visually and has a semantic connotation, such as the \emph{shape} of a hat or the \emph{color} of a ball. Attributes allow characterizing objects in far greater detail than a category label and are therefore the key to several advanced applications, including understanding complex queries in {\em semantic search}, learning about objects from {\em textual description}, and accounting for the content of images in great detail. Textural properties have an important role in object descriptions, particularly for those objects that are best qualified by a pattern, such as a shirt or the wing of bird or a butterfly as illustrated in Fig.~\ref{fig:examples}. Nevertheless, so far the attributes of textures have been investigated only tangentially. In this paper we address the question of whether there exists a ``universal'' set of attributes that can describe a wide range of texture patterns, whether these can be reliably estimated from images, and for what tasks they are useful.

\begin{figure*}[!t]
\raggedright
\foreach \n in {banded, blotchy,braided,bubbly, bumpy,chequered, cobwebbed, cracked, crosshatched, crystalline, dotted, fibrous, flecked, freckled, frilly,gauzy,grid, grooved, honeycombed, interlaced, knitted, lacelike, lined, marbled, matted, meshed, paisley, perforated, pitted, pleated, polka-dotted, porous, potholed, scaly, smeared, spiralled, sprinkled, stained, stratified, striped, studded, swirly, veined, waffled, woven, wrinkled, zigzagged}
{\renewcommand{\baselinestretch}{0.7}\normalsize
\begin{overpic}[width=0.079\textwidth,trim=0 0 -1pt -4pt]{fig/figure2/\n_0002}
\put(0,0){\includegraphics[trim=0 0 130pt -4pt,clip,width=0.039\textwidth]{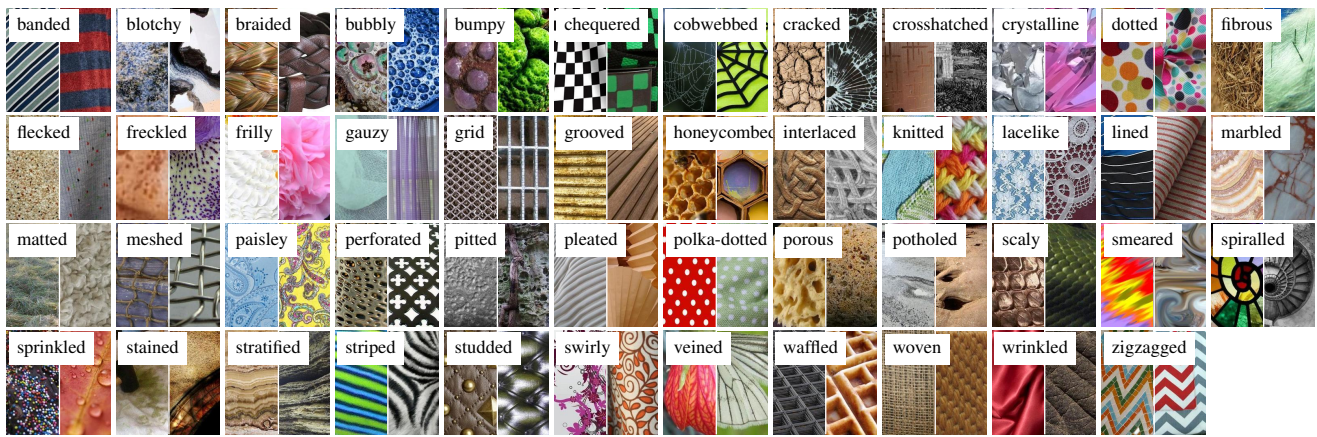}} 
\put(50,0){\linethickness{0.5pt}\color{white}\line(0,1){100}} 
\put(2,78){\scriptsize\colorbox{white}{\strut\n}}
\end{overpic} }
\caption{The 47 texture words in the \textbf{describable texture dataset} introduced in this paper. Two examples of each attribute are shown to illustrate the significant amount of variability in the data.}\label{f:words}
\end{figure*}

The study of perceptual attributes of textures has a long history starting from pre-attentive aspects and grouping~\cite{julesz81textons}, to coarse high-level attributes~\cite{tamura78textural, amadasun89textural,bajcsy73computer}, to some recent work aimed at discovering such attributes by automatically mining descriptions of images from the Internet~\cite{berg2010automatic, ferrari2008learning}. However, the texture attributes investigated so far are rather few or too generic for a detailed description most ``real world'' patterns. Our work is motivated by the one of Bhusan et al.~\cite{bhushan1997texture} who studied the relationship between commonly used English words and the perceptual properties of textures, identifying a set of words sufficient to describing a wide variety of texture patterns. While they study the psychological aspects of texture perception, the focus of this paper is the challenge of estimating such properties from images automatically.

Our {\bf first contribution} is to select a subset of 47 \emph{describable texture attributes}, based on the work of Bhusan et al., that capture a wide variety of visual properties of textures and to introduce a corresponding \emph{describable texture dataset} consisting of 5,640 texture images \emph{jointly} annotated with the 47 attributes (Sect.~\ref{s:dtd}). In an effort to support directly real world applications, and inspired by datasets such as \emph{ImageNet}~\cite{deng09imagenet} and the \emph{Flickr Material Dataset} (FMD)~\cite{sharan09material}, our images are captured ``in the wild'' by downloading them from the Internet rather than collecting them in a laboratory. We also address the practical issue of crowd-sourcing this large set of joint annotations efficiently accounting for the co-occurrence statistics of attributes, the appearance of the textures, and the reliability of annotators (Sect.~\ref{s:design}).

Our {\bf second contribution} is to identify a {\em gold standard texture representation} that achieves optimal recognition of the describable texture attributes in challenging real-world conditions. Texture classification has been widely studied in the context of recognizing materials supported by datasets such as~\emph{CUReT}~\cite{dana99reflectance}, \emph{UIUC}~\cite{lazebnik05sparse}, \emph{UMD}~\cite{xu09viewpoint}, \emph{Outex}~\citep{ojala2002multiresolution}, \emph{Drexel Texture Database}~\cite{oxholm2012texture}, and \emph{KTH-TIPS}~\citep{caputo05class,hayman04learning}. These datasets address material recognition under variable occlusion, viewpoint, and illumination and have motivated the creation of a large number of specialized texture representations that are invariant or robust to these factors~\cite{varma2005statistical,ojala2002multiresolution,varma2003texture,leung2001representing}. In contrast, generic object recognition features such as SIFT was shown to work the best for material recognition in FMD, which, like DTD, was collected ``in the wild''. Our findings are similar, but we also find that Fisher vectors~\cite{perronnin07fisher} computed on SIFT features and certain color features can significantly boost performance. Surprisingly, these descriptors outperform specialized state-of-the-art texture representations not only in recognizing our describable attributes, but also in a variety of datasets for material recognition, achieving an accuracy of 63.3\% on FMD and 67.5\% on KTH-TIPS2-b dataset (Sect.~\ref{s:representation},~\ref{s:exp1}).

Our {\bf third contribution} consists in several \emph{applications} of the proposed describable attributes. These can serve a complimentary role for recognition and description in domains where the material is not-important or is known ahead of time, such as fabrics or wallpapers. However, can these attributes improve other texture analysis tasks such as material recognition?  We answer this question in the affirmative in a series of experiments on the challenging FMD  and KTH datasets. We show that estimates of these properties when used a features can boost recognition rates even more for material classification achieving an accuracy of 53.1\% on FMD and 64.6\% on KTH when used alone as a 47 dimensional feature, and 65.4\% on FMD and 74.6\% on KTH when combined with SIFT and simple color descriptors (Sect.~\ref{s:exp2}). \emph{These represent more than an absolute gain of 8\% in accuracy over previous state of the art. Our 47 dimensional feature contributed with 2.2 to 7\% to the gain}. Furthermore, these attribute are easy to describe by design, hence they can serve as intuitive dimensions to explore large collections of texture patterns -- for e.g., product catalogs (wallpapers or bedding sets) or material datasets. We present several such visualizations in the paper (Sect.~\ref{s:exp3}). 

\section{The describable texture dataset}\label{s:dtd}

This section introduces the \emph{Describable Textures Dataset} (DTD), a collection of real-world texture images annotated with one or more adjectives selected in a vocabulary of 47 English words. These adjectives, or \emph{describable texture attributes}, are illustrated in Fig.~\ref{f:words} and include words such as {\em banded}, \emph{cobwebbed}, \emph{freckled}, \emph{knitted}, and \emph{zigzagged}.

DTD investigates the problem of {\bf texture description}, intended as the recognition of describable texture attributes. This problem differs from the one of {\em material recognition} considered in existing datasets such as CUReT, KTH, and FMD. While describable attributes are correlated with materials, attributes do not imply materials (\eg \emph{veined} may equally apply to leaves or marble) and materials do not imply attributes (not all marbles are \emph{veined}). Describable attributes can be {\em combined} to create rich descriptions (Fig.~\ref{f:co-occurence}; marble can be \emph{veined}, \emph{stratified} and \emph{cracked} at the same time), whereas a typical assumption is that textures are made of a single material. Describable attributes are \emph{subjective} properties that depend on the imaged object as well as on human judgments, whereas materials are objective. In short, attributes capture properties of textures {\em beyond} materials, supporting human-centric tasks where describing textures is important. At the same time, they will be shown to be helpful in material recognition as well (Sect.~\ref{s:high} and~\ref{s:exp2}).

DTD contains {\bf textures in the wild}, \ie texture images extracted from the web rather than begin captured or generated in a controlled setting. Textures fill the images, so we can study the problem of texture description independently of texture segmentation. With 5,640 such images, this dataset aims at supporting real-world applications were the recognition of texture properties is a key component. Collecting images from the Internet is a common approach in categorization and object recognition, and was adopted in material recognition in FMD. This choice trades-off the systematic sampling of illumination and viewpoint variations existing in datasets such as CUReT, KTH-TIPS, Outex, and Drexel datasets for a representation of real-world variations, shortening the gap with applications. Furthermore, the invariance of describable attributes is not an intrinsic property as for materials, but it reflects invariance in the human judgments, which should be captured empirically.

DTD is designed as a {\bf public benchmark}, following the standard practice of providing 10 preset splits into equally-sized training, validation and test subsets for easier algorithm comparison (these splits are used in all the experiments in the paper). DTD will be made publicly available on the web at [annonymized], along with standardized evaluation, as well as code reproducing the results in Sect.~\ref{s:experiments}.

\paragraph{Related work.} Apart from material datasets, there have been numerous attempts at collecting attributes of textures at a smaller scale, or in controlled settings. Our work is related to the work of~\cite{matthews13enriching}, where they analyzed images in the Outex dataset~\citep{ojala2002multiresolution} using a subset of the attributes we consider. Their attributes were demonstrated to perform better than several low-level descriptors, but these were trained and evaluated on the \emph{same} dataset. Hence it is not clear if their learned attributes generalize well to other settings. In contrast, we show that: (i) our texture attributes trained on DTD outperform their semantic attributes on Outex and (ii) they can significantly boost performance on a number of other material and texture benchmarks (Sect.~\ref{s:exp2}).

\subsection{Dataset design and collection}\label{s:design}

This section discusses how DTD was designed and collected, including: selecting the 47 attributes, finding at least 120 representative images for each attribute, collecting a full set of multiple attribute labels for each image in the dataset, and addressing annotation noise.

\paragraph{Selecting the describable attributes.} Psychological experiments suggest that, while there are a few hundred words that people commonly use to describe textures, this vocabulary is redundant and can be reduced to a much smaller number of representative words. Our starting point is the list of $98$ words identified by Bhusan, Rao and Lohse~\cite{bhushan1997texture}. Their seminal work aimed to achieve for texture recognition the same that color words have achieved for describing color spaces~\cite{berlin1991basic}. However, their work mainly focuses on the cognitive aspects of texture perception, including perceptual similarity and the identification of  directions of perceptual texture variability. Since we are interested in the visual aspects of texture, we ignored words such as ``corrugated" that are more related to surface shape properties, and  words such as ``messy" that do not necessarily correspond to visual features. After this screening phase we analyzed the remaining words and merged similar ones such as ``coiled", ``spiraled"  and ``corkscrewed" into a single term. This resulted in a set of $47$ words, illustrated in Fig.~\ref{f:words}.

\paragraph{Bootstrapping the key images.} Given the 47 attributes, the next step was collecting a sufficient number (120) of example images representative of each attribute. A very large initial pool of about a hundred-thousands images was downloaded from Google and Flickr by entering the attributes and related terms as search queries. Then Amazon Mechanical Turk (AMT) was used to remove low resolution, poor quality, watermarked images, or images that were not almost entirely filled with a texture. Next, detailed annotation instructions were created for each of the 47 attributes, including a dictionary definition of each concept and examples of correct and incorrect matches. Votes from three AMT annotators were collected for the candidate images of each attribute and a shortlist of about $200$ highly-voted images was further manually checked by the authors to eliminate residual errors. The result was a selection of $120$ {\em key representative images} for each attribute.

\begin{figure*}[t]
\includegraphics[width=0.70\textwidth]{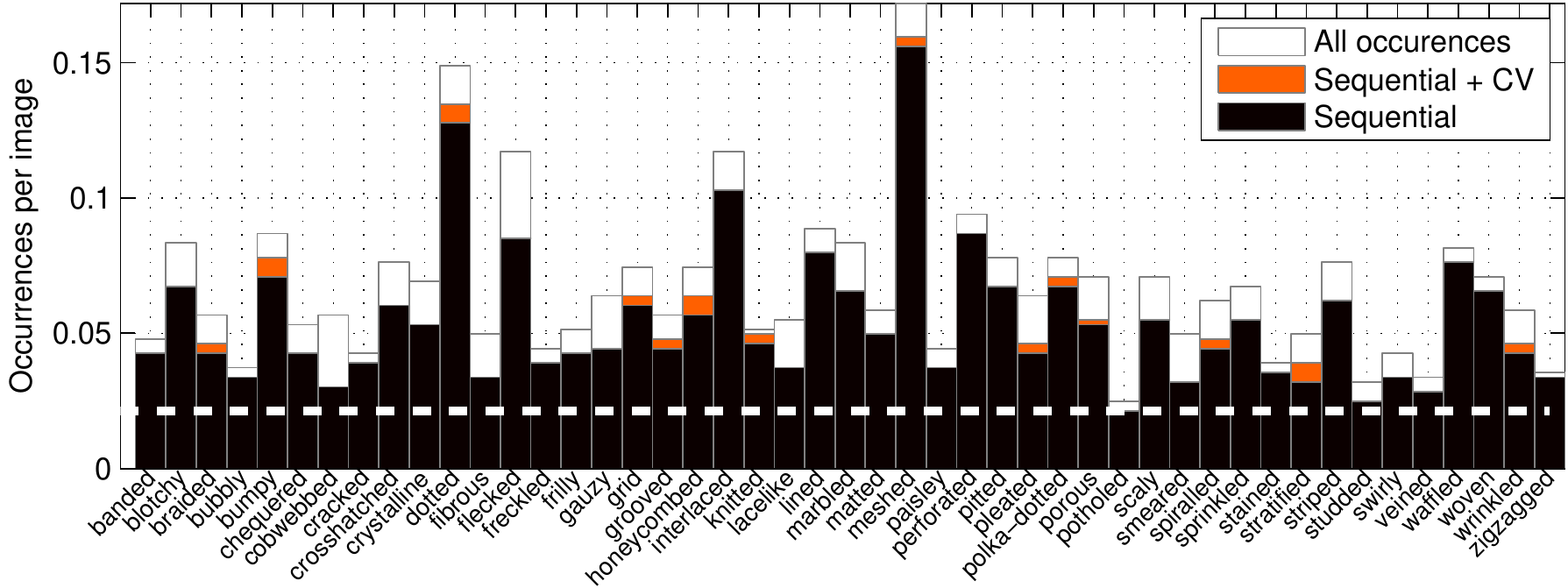}\hfill
\raisebox{0.85in}{
\setlength{\tabcolsep}{3pt}
\begin{tabular}{cc}
\raisebox{0.8in}{$q$} & {\setlength\fboxsep{0pt}\fbox{\includegraphics[width=0.25\textwidth]{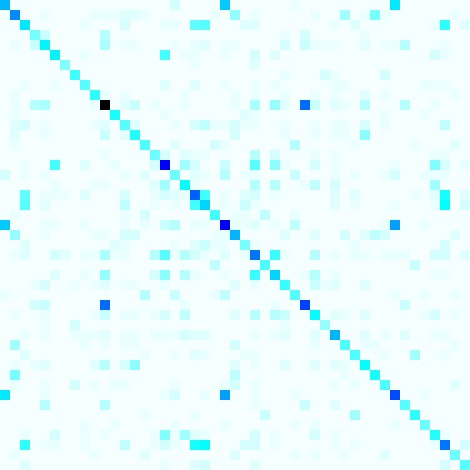}}}\\
& $q'$ \\
\end{tabular}
}
\caption{{\bf Quality of joint sequential annotations.} Each bar shows the average number of occurrences of a given attribute in a DTD image. The horizontal dashed line corresponds to a frequency of 1/47, the minimum given the design of DTD (Sect.~\ref{s:design}). The black portion of each bar is the amount of attributes discovered by the sequential procedure, using only 10 annotations per image (about one fifth of the effort required for exhaustive annotation). The orange portion shows the additional recall obtained by integrating CV in the process. {\bf Right: co-occurrence of attributes.} The matrix shows the joint probability $p(q,q')$ of two attributes occurring together (rows and columns are sorted in the same way as the left image).}\label{f:ja}\label{f:co-occurence}
\end{figure*}

\paragraph{Sequential join annotations.} So far only the key attribute of each texture image is known while any of the remaining 46 attributes may apply as well. Exhaustively collecting annotations for 46 attributes and 5,640 texture images was found to be too expensive. To reduce this cost we propose to exploiting the correlation and sparsity of the attribute occurrences (Fig.~\ref{f:co-occurence}). For each attribute $q$, twelve key images are annotated exhaustively and used to estimate the probability $p(q'|q)$ that {\em another} attribute $q'$ could co-exist with $q$. Then for the remaining key images of attribute $q$, only annotations for attributes $q'$ with non negligible probability -- in practice 4 or 5 -- are collected, assuming that the attributes would not apply. This procedure occasionally misses attribute annotations; Fig.~\ref{f:ja} evaluates attribute recall by 12-fold cross-validation on the 12 exhaustive annotations for a fixed budget of collecting 10 annotations per image (instead of 47).

A further refinement is to suggest which attributes $q'$ to annotated not just based on $q$, but also based on the individual appearance of an image $\ell_i$. This was done by using the attribute classifier learned in Sect.~\ref{s:experiments}; after  Platt's calibration~\cite{platt00probabilistic} on an held-out test set, the classifier score $c_{q'}(\ell_i) \in \real$ is transformed in a probability
$p(q'|\ell_i) = \sigma(c_{q'}(\ell))$ where  $\sigma(z) = 1 / (1 + e^{-z})$ is the sigmoid function. By construction, Platt's calibration reflects the prior probability $p(q') \approx p_0 = 1/47$ of $q'$ on the validation set. To reflect the probability $p(q'|q)$ instead, the score is adjusted as
\[
 p(q'|\ell_i, q) \propto
 \sigma(c_{q'}(\ell_i)) \times
 \frac{p(q'|q)} {1 -  p(q'|q)} \times
 \frac{1 - p_0}{p_0}
\]
and used to find which attributes to annotated for each image. As shown in Fig.~\ref{f:ja}, for a fixed annotation budged this method increases attribute recall. Overall, with roughly 10 annotations per images it was possible to recover of all the attributes for at least 75\% of the images, and miss one out of four (on average) for another 20\% while keeping the annotation cost to a reasonable level.

\paragraph{Handling noisy annotations.} So far it was assumed that annotators are perfect: deterministic and noise-free. This is not the case, in part due to the intrinsic subjectivity of describable texture attributes, and in part due to distracted, adversarial, or unqualified annotators. As commonly done, we address this problem by collecting the same annotation multiple times (five) using different annotators, and forming a consensus.

Beyond simple voting, we found that the method of~\cite{welinder10online} can effectively remove or down-weigh bad annotators improving agreement. This method models each annotator $\alpha_j$ as a classifier with a given bias and error rate. Then, given a collection $\hat{a}_{qij}\in\{0,1\}$ of binary annotations for attribute $q$, image $i$, and annotator $j$, it tries to estimate simultaneously the ground truth labels $a_{qi}$ and the quality $\alpha_j$ of the individual annotators. The method is appealing as several quantities are easily interpretable. For example, the prior $p(\alpha_j)$ on annotators encodes how frequently we expect to find good and bad annotators (\eg we found that 0.5\% of them labeled images randomly). A major difference compared to the scenario considered in~\cite{welinder10online} is that, in our case, the key attribute of each image is already known. By incorporating this as additional prior, the method can use the key attributes to implicitly benchmark and calibrate annotators. The final set of annotations $\{a_{qi}\}$ is obtained by thresholding the (approximated) posterior marginal $p(a_{qi}|\{\hat a_{qij}\})$ to 60\%, similar to choosing three out of five votes in the basic voting scheme, computed using variational inference. In general, we found most probabilities to be very close to 100\% or 0\%, suggesting that there is little residual noise in the process. We also inspected the top 30 images of each attribute based on simple voting and this posterior marginals and found the ranking to be significantly improved.

\section{Texture representations}\label{s:representation}

Given the DTD dataset developed in Sect.~\ref{s:dtd}, this section moves on to the problem of designing a system that can automatically recognize the attributes of textures. Given a texture image $\ell$ the first step is to compute a {\em representation} $\phi(\ell) \in \real^d$ of the image; the second step is to use a classifier such as a Support Vector Machine (SVM) $\langle \bw, \phi(\ell)\rangle$ to score how strongly the $\ell$ matches a given perceptual category. We propose two such representations: a gold-standard low-level texture descriptor based on the improved Fisher Vector (Sect.~\ref{s:low}) and a mid-level texture descriptor consisting of the describable attributes themselves (Sect.~\ref{s:high}). The details of the classifiers are discussed in Sect.~\ref{s:experiments}.

\subsection{Improved Fisher vectors}\label{s:low}

This section introduces our gold-standard low-level texture representation, the {\em Improved Fisher Vector} (IFV) of and relates it to existing texture descriptors. We port IFV from the object recognition literature~\cite{perronnin10improving} and we show that it substantially outperforms specialized texture representations (Sect.~\ref{s:experiments}).

Given an image $\ell$, the {\em Fisher Vector} (FV) formulation of~\cite{perronnin07fisher} starts by extracting local SIFT~\cite{lowe99object} descriptors $\{\bd_1,\dots,\bd_n\}$ densely and at multiple scales. It then soft-quantizes the descriptors by using a Gaussian Mixture Model (GMM) with $K$ modes, prior probabilities $\pi_k$, mode means $\mu_k$ and mode covariances $\Sigma_k$. Covariance matrices are assumed to be diagonal, but local descriptors are first decorrelated and optionally dimensionality reduced by PCA. Then first and second order statistics are computed as
\begin{align*}
u_{jk} &=
{1 \over {n \sqrt{\pi_k}}}
\sum_{i=1}^{n}
q_{ik} \frac{d_{ji} - \mu_{jk}}{\sigma_{jk}},
\\
v_{jk} &=
{1 \over {n \sqrt{2 \pi_k}}}
\sum_{i=1}^{n}
q_{ik} \left[ \left(\frac{d_{ji} - \mu_{jk}}{\sigma_{jk}}\right)^2 - 1 \right],
\end{align*}
where $j$ spans descriptor dimensions and $q_{ik}$ is the posterior probability of mode $k$ given descriptor $\bd_i$, i.e. $q_{ik} \propto \exp\left[-\frac{1}{2} (\bd_i - \mu_k)^T \Sigma_k^{-1} (\bd_i - \mu_k)\right]$. These statistics are then stacked into a vector $(\bu_1, \bv_1, \dots, \bu_K, \bv_K)$. In order to obtain the {\em improved} version of the representation, the signed square root $\sqrt{|z|} \operatorname{sign} z$ is applied to its components and the vector is $l^2$ normalized.

At least two key ideas in IFV were pioneered in texture analysis: the idea of sum-pooling local descriptors was introduced by~\cite{malik90preattentive}, and the idea of quantizing local descriptors to construct histogram of features was pioneered by~\cite{leung2001representing} with their computational model of textons. However, three key aspects of the IFV representation were developed in the context of object recognition. The first one is the use of the SIFT descriptors, originally developed for object matching~\cite{lowe99object}, that are more distinctive that local descriptors popular in texture analysis such as filter banks~\cite{leung2001representing,varma2005statistical,geusebroek03fast}, local intensity patterns~\cite{ojala2002multiresolution}, and patches~\cite{varma2003texture}. The second one is replacing histogramming with the more expressive FV pooling method ~\cite{perronnin07fisher}. And the third one is the use of the square-root kernel map~\cite{perronnin10improving} in the improved version of the Fisher Vector.

We are not the first to use SIFT or IFV in texture recognition. For example, SIFT was used in~\cite{sharan13recognizing}, and Fisher Vectors were used in~\cite{sharma12local}. However, neither work tested the standard IFV formulation~\cite{perronnin10improving}, which is well tuned for object recognition, developing instead variations specialized for texture analysis. We were therefore somewhat surprised to discover that the off-the-shelf method surpasses these approaches (Sect.~\ref{s:exp1}).

\subsection{Describable attributes as a representation}\label{s:high}

The main motivation for recognizing describable attributes is to support human-centric applications, enriching the vocabulary of visual properties that machines can understand. However, once extracted, these attributes may also be used as texture descriptors in their own right. As a simple incarnation of this idea, we propose to collect the response of attribute classifiers trained on DTD in a 47-dimensional feature vector $\phi(\ell) = (c_1(\ell), \dots, c_{47}(\ell))$. Sect.~\ref{s:experiments} shows that this very compact representation achieves excellent performance in material recognition; in particular, combined with IFV (SIFT and color) it sets the new state-of-the-art on KTH-TIPS2-b and FMD. In addition to the contribution to the best results, our proposed attributes generate meaningful descriptions of the materials from KTH-TIPS2-b (aluminium foil: wrinkled; bread: porous).

\section{Experiments}\label{s:experiments}

\subsection{Improved Fisher Vectors for textures}\label{s:exp1}

This section demonstrates the power of IFV as a texture representation by comparing it to established texture descriptors. Most of these representations can be broken down into two parts: computing local image descriptors $\{\bd_1,\dots,\bd_n\}$ and encoding them into a global image statistics $\phi(\ell)$.

In IFV the {\bf local descriptors} $\bd_i$ are 128-dimensional {\em SIFT} features, capturing a spatial histogram of the local gradient orientations; here spatial bins have an extent of $6 \times 6$ pixels and descriptors are sampled every two pixels and at scales $2^{i/3},i=0,1,2,\dots$. We also evaluate as local descriptors the \emph{Leung and Malik} (LM)~\cite{leung2001representing} (48-D) and \emph{MR8} (8-D)~\cite{varma2005statistical,geusebroek03fast} filter banks, the $3 \times 3$ and $7\times 7$ raw image patches of\cite{varma2003texture}, and the {\em local binary patterns} (LBP) of~\cite{ojala2002multiresolution}.

{\bf Encoding} maps image descriptors $\{\bd_1,\dots,\bd_n\}$ to a statistics $\phi(\ell) \in \real^d$ suitable for classification. Encoding can be as simple as averaging (sum-pooling) descriptors~\cite{malik90preattentive}, although this is often preceded by a high-dimensional sparse coding step. The most common coding method is to vector quantize the descriptors using an algorithm such as $K$-means~\cite{leung2001representing}, resulting in the so-called {\em bag-of-visual-words} (BoVW) representation~\cite{csurka04visual}. Variations include soft quantization by a GMM in FV (Sect.~\ref{s:low}) or specialized quantization schemes, such as mapping LBPs to {\em uniform patterns}~\cite{ojala2002multiresolution} (LBP$^{u}$; we use the rotation invariant multiple-radii version of~\cite{matthews13enriching} for comparison purposes). For LBP, we also experiment with a variant (LBP-VQ) where standard LBP$^{u2}$ is computed in $8 \times 8$ pixel neighborhoods, and the resulting local descriptors are further vector quantized using $K$-means and pooled as this scheme performs significantly better in our experiments.

For each of the selected features, we experimented with several {\bf SVM kernels}: linear $K(\bx',\bx'') = \langle \bx', \bx''\rangle$, Hellinger's $\sum_{i=1}^d \sqrt{x_i' x_i''}$, additive-$\chi^2$ $\sum_{i=1}^d {x_i' x_i''}/(x_i'+x_i'')$, and exponential-$\chi^2$ $\exp\left[ - \lambda \sum_{i=1}^d {(x_i' -x_i'')^2}/(x_i'+x_i'') \right]$ kernels sign-extended as in~\cite{vedaldi10efficient}. In the latter case, $\lambda$ is selected as one over the mean of the kernel matrix on the training set. The data is normalized so that $K(\bx',\bx'')=1$ as this is often found to improve performance. Learning uses a standard non-linear SVM solver and validation in order to select the parameter $C$ in the range $\{0.1, 1, 10, 100\}$ (the choice of $C$ was found to have little impact on the result).

\begin{table}
\centering
{\small
\setlength{\tabcolsep}{3pt}
\begin{tabular}{l|cccc}
                     & \multicolumn{4}{|c}{{Kernel}}                                                                       \\ \hline
{Local d.}           & {Linear}                & {Hellinger}             & add-$\chi^2$            & exp-$\chi^2$          \\ \hline
{MR8}                & 15.9  $\pm$ 0.8         & 19.7 $\pm$ 0.8          & 24.1 $\pm$ 0.7          & 30.7 $\pm$ 0.7        \\
{LM}                 & 18.8  $\pm$  0.5        & 25.8 $\pm$ 0.8          & 31.6 $\pm$ 1.1          & 39.7 $\pm$ 1.1        \\
{Patch$_{3\times3}$} & 14.6  $\pm$  0.6        & 22.3 $\pm$ 0.7          & 26.0 $\pm$ 0.8          & 30.7 $\pm$ 0.9        \\
{Patch$_{7\times7}$} & 18.0  $\pm$  0.4        & 26.8 $\pm$ 0.7          & 31.6 $\pm$ 0.8          & 37.1 $\pm$ 1.0        \\
{LBP$^{u}$}         & 8.2  $\pm$  0.4         & 9.4 $\pm$ 0.4           & 14.2 $\pm$ 0.6          & 24.8 $\pm$ 1.0        \\
{LBP-VQ}             & 21.1  $\pm$  0.8        & 23.1 $\pm$ 1.0          & 28.5 $\pm$ 1.0          & 34.7 $\pm$ 1.3        \\
{SIFT}               & \textbf{34.7 $\pm$ 0.8} & \textbf{45.5 $\pm$ 0.9} & \textbf{49.7 $\pm$ 0.8} & \textbf{53.8 $\pm$ 0.8} \\
\end{tabular}
}
\caption{Comparison of local descriptors and kernels on the DTD data, averaged over ten splits.}
\label{tbl:results}
\end{table}

\begin{table*}[t]
\centering
{\small
\setlength{\tabcolsep}{3pt}
\begin{tabular}{l|ccc|c|ccc}
\multirow{2}{*}{Dataset} & \multicolumn{3}{|c|}{SIFT} & \multicolumn{2}{|c}{Published}                                                                                 \\
                         & IFV                       & BoVW           & VLAD           & Best                                               & \citep{sifre13rotation} \\ \hline
{CUReT}                  & \textbf{99.6 $\pm$ 0.3}   & 98.1 $\pm$ 0.9 & 98.8 $\pm$ 0.6 & $\rightarrow$                                      & 99.4                    \\
{UMD}                    & 99.2 $\pm$ 0.4            & 98.1 $\pm$ 0.8 & 99.3 $\pm$ 0.4 & $\rightarrow$                                      & \textbf{99.7 $\pm$ 0.3} \\
{UIUC}                   & 97.0 $\pm$ 0.9            & 96.1 $\pm$ 2.4 & 96.5 $\pm$ 1.0 & $\rightarrow$                                      & \textbf{99.4 $\pm$ 0.4} \\
{KTH-TIPS}               & \textbf{99.7 $\pm$ 0.1}   & 98.6 $\pm$ 1.0 & 99.2 $\pm$ 0.8 & $\rightarrow$                                      & 99.4  $\pm$ 0.4         \\
{KTH-TIPS-2a}$^\alpha$   & \textbf{82.5 $\pm$ 5.2}   & \textbf{74.8 $\pm$ 5.4} & \textbf{76.5 $\pm$ 5.2} & 73.0 $\pm$ 4.7 \citep{sharma12local}               & --                      \\
{KTH-TIPS-2b}$^{\beta}$  & \textbf{69.3 $\pm$ 1.0}   & 58.4 $\pm$ 2.2 & 63.1 $\pm$ 2.1 & 66.3 \citep{timofte12trainingfree}                 & --                      \\
{FMD}                    & \textbf{58.2 $\pm$ 1.7}   & 49.5 $\pm$ 1.9 & 52.6 $\pm$ 1.5 & 57.1 / 55.6~\citep{sharan13recognizing}$^{\gamma}$ & $41.4 \pm 1.3$          \\ \hline
\textbf{{DTD}}           & \textbf{61.5 $\pm$ 1.4}   & 55.6 $\pm$ 1.3 & 59.8 $\pm$ 1.0 & --                                                 & $40.2 \pm 0.5$          \\
\end{tabular}
\vline
\vline
\vline
\hfill 
\begin{tabular}{ l|c|c }
  Feature & KTH-TIPS-2b & FMD \\
  \hline
  \DTDLIN & 61.1 $\pm$ 2.8 & 48.9 $\pm$ 1.9 \\
  \DTDRBF & 64.6 $\pm$ 1.5 & 53.1 $\pm$ 2.0 \\
  \hline
  \IFVSIFT & 69.3 $\pm$ 1.0 & 58.2$\pm$ 1.7 \\
  \IFVRGB & 58.8 $\pm$ 2.5 & 47.0 $\pm$ 2.7 \\
  \IFVSIFT + \IFVRGB & 67.5 $\pm$ 3.3 & 63.3 $\pm$ 1.9 \\
  \hline
  \DTDRBF + \IFVSIFT & 68.4 $\pm$ 1.4 & 60.1 $\pm$ 1.6 \\
  \DTDRBF + \IFVRGB  & 70.9 $\pm$ 3.5 & 61.3 $\pm$ 2.0  \\
  All three & \textbf{74.6 $\pm$ 3.0} & \textbf{65.4 $\pm$ 2.0} \\
  \hline
  Prev. state of the art & 66.3~\cite{timofte12trainingfree} & 57.1~\cite{sharan13recognizing} \\
\end{tabular}

}
\caption{{\bf Left:} Comparison of encodings and state-of-the-art texture recognition methods on DTD as  well as standard material recognition benchmarks. $\alpha:$ three samples for training, one for evaluation; $\beta:$ one sample for training, three for evaluation. $\gamma:$ with/without ground truth masks (\cite{sharan13recognizing} Sect. 6.5); our results do not use them. {\bf Right:} Combined with \IFVSIFT and \IFVRGB, the \DTDRBF features achieve a significant improvement in classification performance on the challenging KTH-TIPS-2b and FMD compared to published state of the art results.}
\label{tbl:dataset-results}\label{tbl:fmd-results}
\end{table*}

\begin{figure}[t]
\centering
\includegraphics[width=\linewidth]{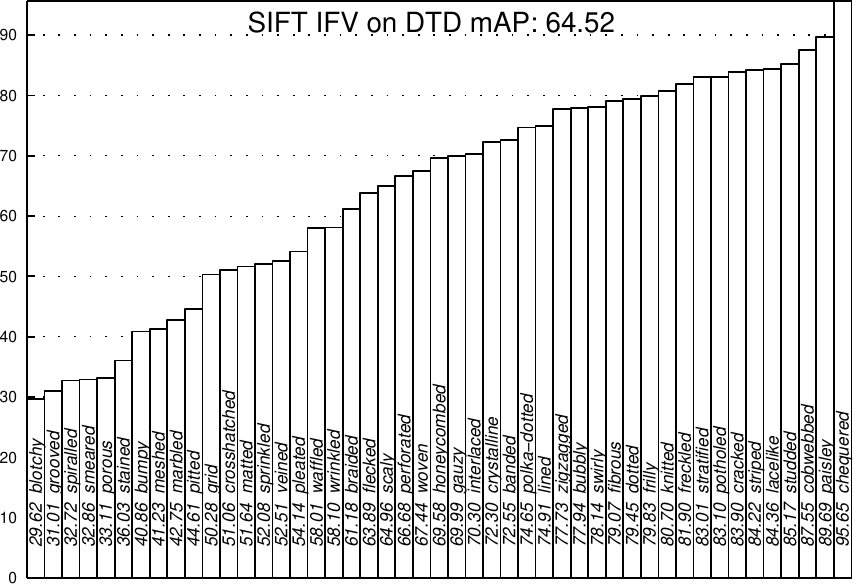}
\caption{Per-class AP of the 47 describable attribute classifiers on DTD using the \IFVSIFT representation and linear classifiers.}\label{fig:perclassap}
\end{figure}

\paragraph{Local descriptor comparisons on DTD.} This experiments compares local descriptors and kernels on DTD. All comparison use the bag-of-visual-word pooling/encoding scheme using $K$-means for vector quantization the descriptors. The DTD data is used as a benchmark averaging the results on the ten train-val-test splits. $K$ was cross-validated, finding an optimal setting of 1024 visual words for SIFT and color patches, 512 for LBP-VQ, 470 for the filter banks. Tab.~\ref{tbl:results}, reports the mean Average Precision (mAP) for 47 SVM attribute classifiers. As expected, the best kernel is exp-$\chi^2$, followed by additive $\chi^2$ and Hellinger, and then linear. Dense SIFT (53.82\% mAP) outperforms the best specialized texture descriptor on the DTD data (39.67\% mAP for LM). Fig.~\ref{fig:perclassap} shows AP for each attribute: concepts like \emph{chequered} achieve nearly perfect classification, while others such as \emph{blotchy} and \emph{smeared} are far harder.

\paragraph{Encoding comparisons on DTD.} This experiment compares three encodings: BoVW, VLAD~\citep{jegou10aggregating} and IFV. VLAD is similar to IFV, but uses $K$-means for quantization and stores only first-order statistics of the descriptors. Dense SIFT is used as a baseline descriptor and performance is evaluated on ten splits of DTD in Tab.~\ref{tbl:dataset-results}. IFV (256 Gaussian modes) and VLAD (512 $K$-means centers) performs similarly (about 60\% mAP) and significantly better than BoVW (53.82\% mAP). As we will see next, however, IFV significantly outperforms VLAD in other texture datasets. We also experimented with the state-of-the-art descriptor of \citep{sifre13rotation} which we did not find to be competitive with IFV on FMD and DTD (Tab.~\ref{tbl:dataset-results}); unfortunately could not obtain an implementation of~\citep{sharan13recognizing} to try on our data -- however \IFVSIFT outperforms it on material recognition. 

\paragraph{State-of-the-art material classification.} This experiments evaluates the encodings on several material recognition datasets: CUReT~\cite{dana99reflectance}, UMD~\cite{xu09viewpoint}, UIUC~\cite{lazebnik05sparse}, KTH-TIPS~\cite{hayman04learning}, KTH-TIPS2(a and b)~\cite{caputo05class}, and FMD~\cite{sharan09material}. Tab.~\ref{tbl:dataset-results} compares with the existing state-of-the-art~\citep{timofte12trainingfree,sifre13rotation,sharma12local} on each of them. For saturated datasets such as CUReT, UMD, UIUC, KTH-TIPS the performance of most methods is above to 99\% mean accuracy and there is little difference between them. In harder datasets the advantage of IFV is evident: KTH-TIPS-2a (+5\%), KTH-TIPS-2b (+3\%), and FMD (+1\%). In particular, while FMD includes manual segmentations of the textures, these are not used here here. Furthermore, IFV is conceptually simpler than the multiple specialized features used in~\citep{sharma12local} for material recognition.

\subsection{Describable attributes as a representation}\label{s:exp2}

This section evaluates using the 47 describable attributes as a texture descriptor applying it to the task of material recognition. The attribute classifiers are trained on DTD using the IFV+SIFT representation and linear classifiers as in the previous section (\DTDLIN). As explained in Sect.~\ref{s:high}, these are then used to form 47-dimensional descriptors of each texture image in FMD and KTH-TIPS2-b.

When combined with a linear SVM classifier, results are promising (Tab.~\ref{tbl:fmd-results}): on KTH-TIPS2-b, the describable attributes yield 61.1\% mean accuracy and
49.0\% on FMD outperforming the aLDA model of~\citep{sharan13recognizing} combining color, SIFT and edge-slice (44.6\%). While results are not as good as the \IFVSIFT representation, the dimensionality of this descriptor is \emph{three orders of magnitude smaller} than IFV. For this reason, using an RBF classifier with the DTD features is relatively cheap. Doing so improves the performance by 3.5--4\% (\DTDRBF).

We also investigated combining multiple features: \DTDRBF with \IFVSIFT and \IFVRGB. \IFVRGB computes the IFV representation on top of all the $3\times 3$ RGB patches in the image in the spirit of~\cite{varma2003texture}. The performance of \IFVRGB is notable given the simplicity of the local descriptors; however, it is not as good as \DTDRBF which is also 26 times smaller. The combination of \IFVSIFT and \IFVRGB is already notably better than the previous state-of-the-art results and the addition of \DTDRBF improves by another significant margin. Overall, our best result on KTH-TIPS-2b is \textbf{74.6\%} (vs. the previous best of 66.3) and on FMD of \textbf{65.4\%} (vs. 57.1) on FMD, with an improvement of more than~\textbf{8\%} accuracy in both cases.

Finally, we compared the semantic attributes of~\cite{matthews13enriching} with \DTDLIN on the Outex data. Using \IFVSIFT as an underlying representation for our attributes, we obtain 49.82\% mAP on the retrieval experiment of~\cite{matthews13enriching}, which is is not as good as their result with $\text{LBP}^u$ (63.3\%). However, $\text{LBP}^u$ was developed on the Outex data, and it is therefore not surprising that it works so well. To verify this, we retrained our DTD attributes with IFV using $\text{LBP}^u$ as local descriptor, obtaining a score of 64.52\% mAP. This is remarkable considering that their retrieval experiment contains the data used to \emph{train} their own attributes (target set), while our attributes are trained on a completely different data source. Tab.~\ref{tbl:results} shows that $\text{LBP}^u$ is not competitive on DTD.

\renewcommand{\dim}{.09\textwidth}
\begin{figure*}[!t]
\vspace{-0.1in}
\centering
	\subfloat[aluminium]{\includegraphics[width=\dim]{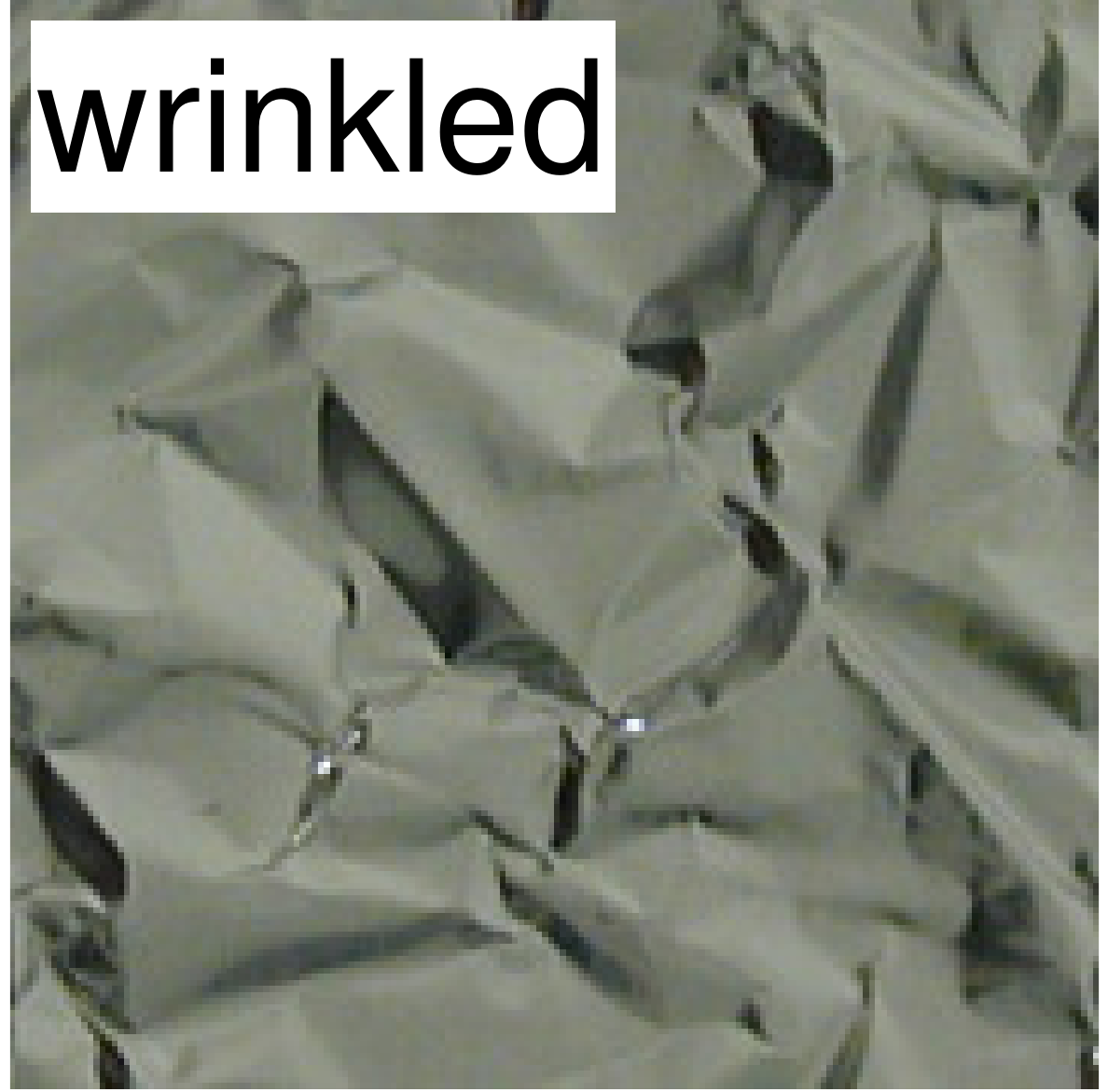}}
	\hfill
	\subfloat[brown bread]{\includegraphics[width=\dim]{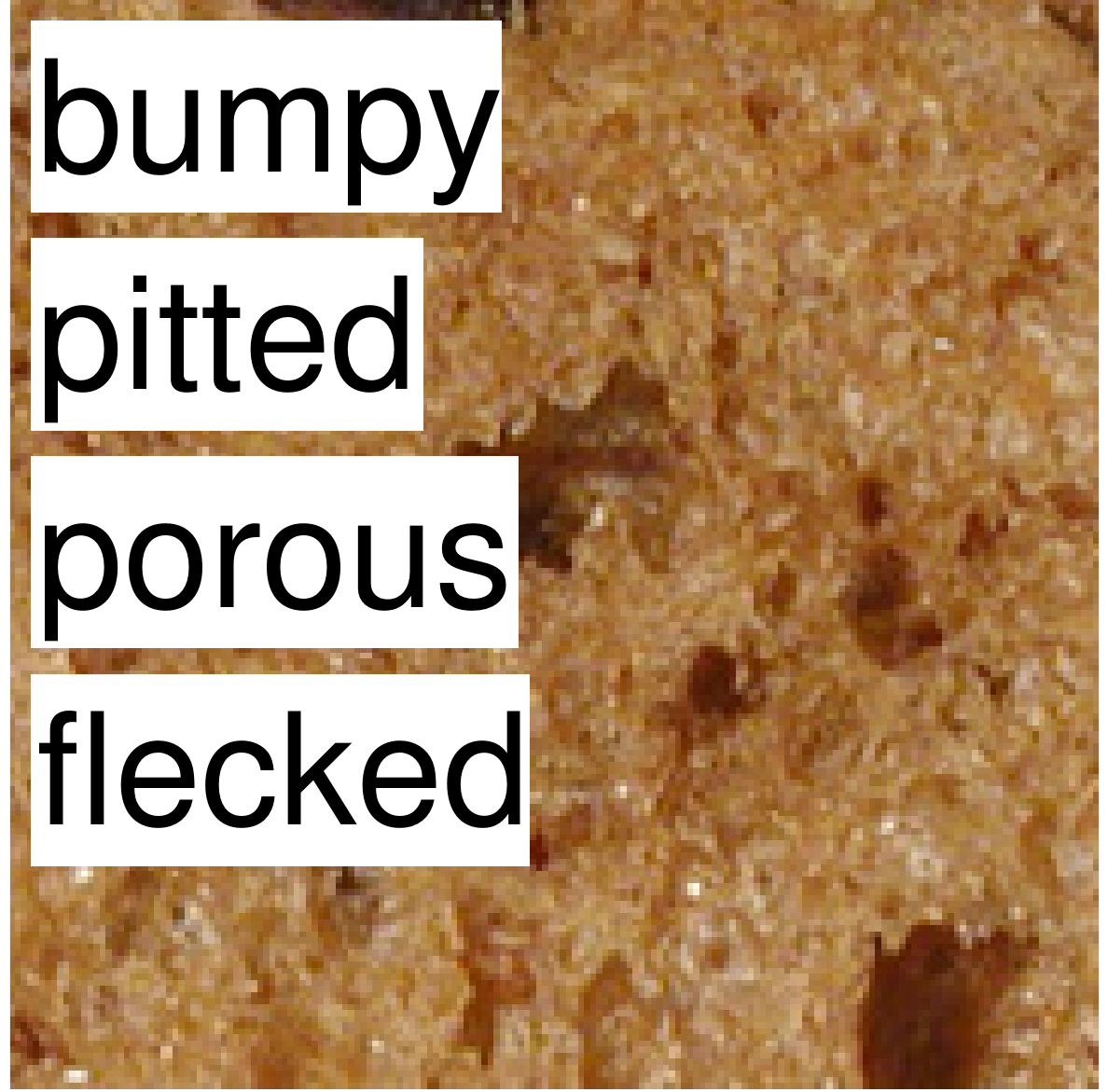}}
	\hfill
	\subfloat[corduroy]{\includegraphics[width=\dim]{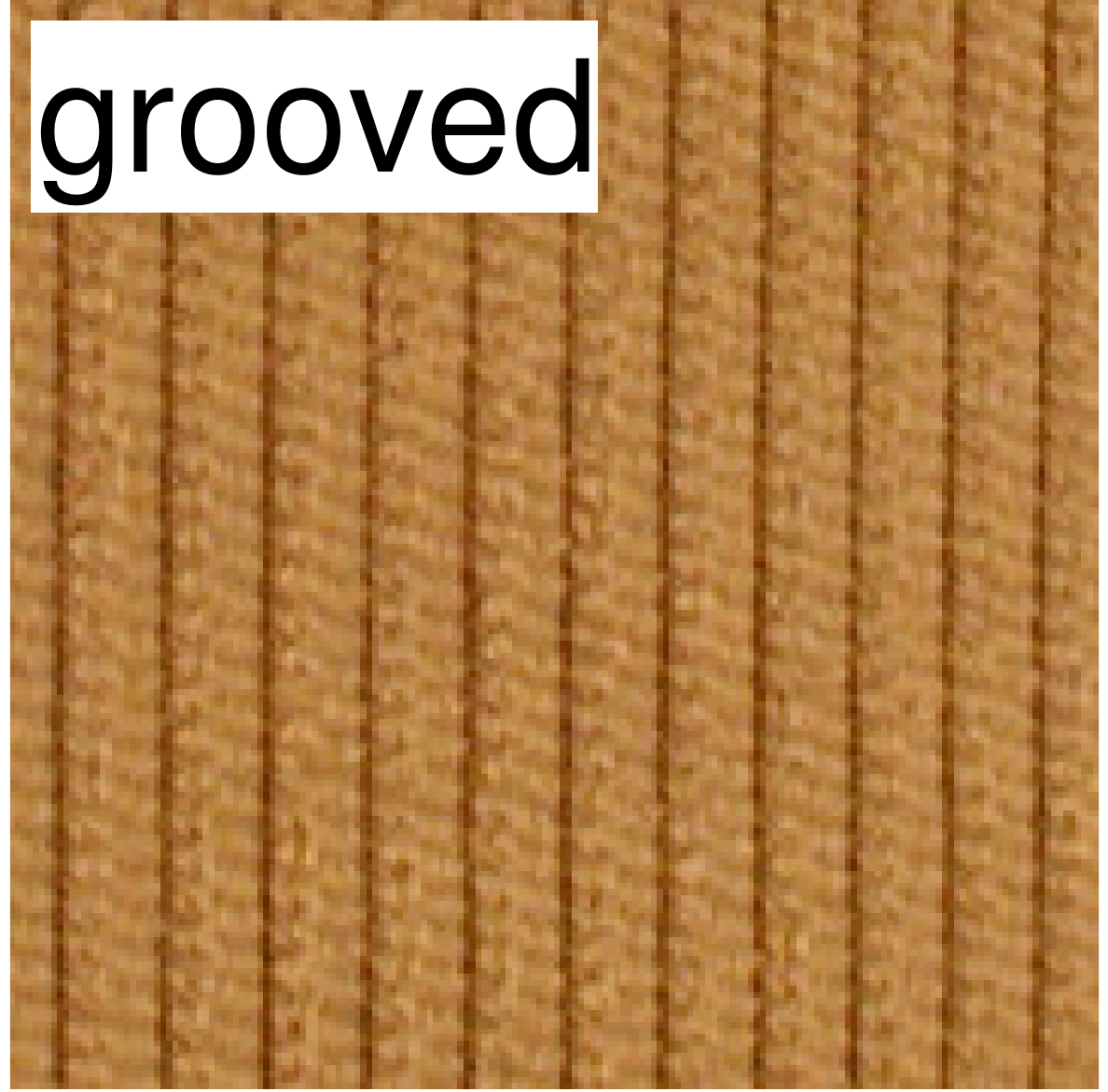}}
	\hfill
	\subfloat[cork]{\includegraphics[width=\dim]{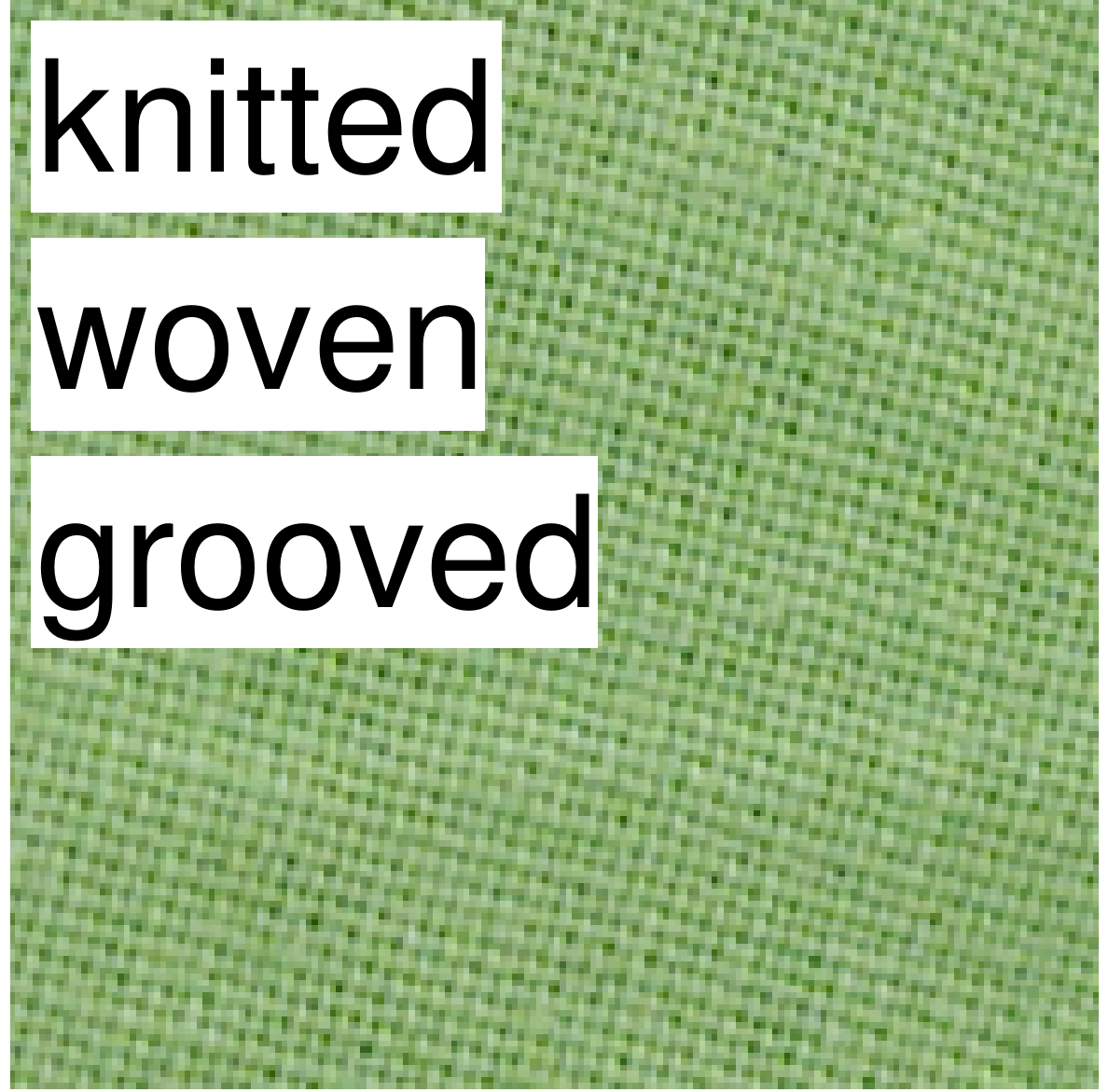}}
	\hfill
	\subfloat[cotton]{\includegraphics[width=\dim]{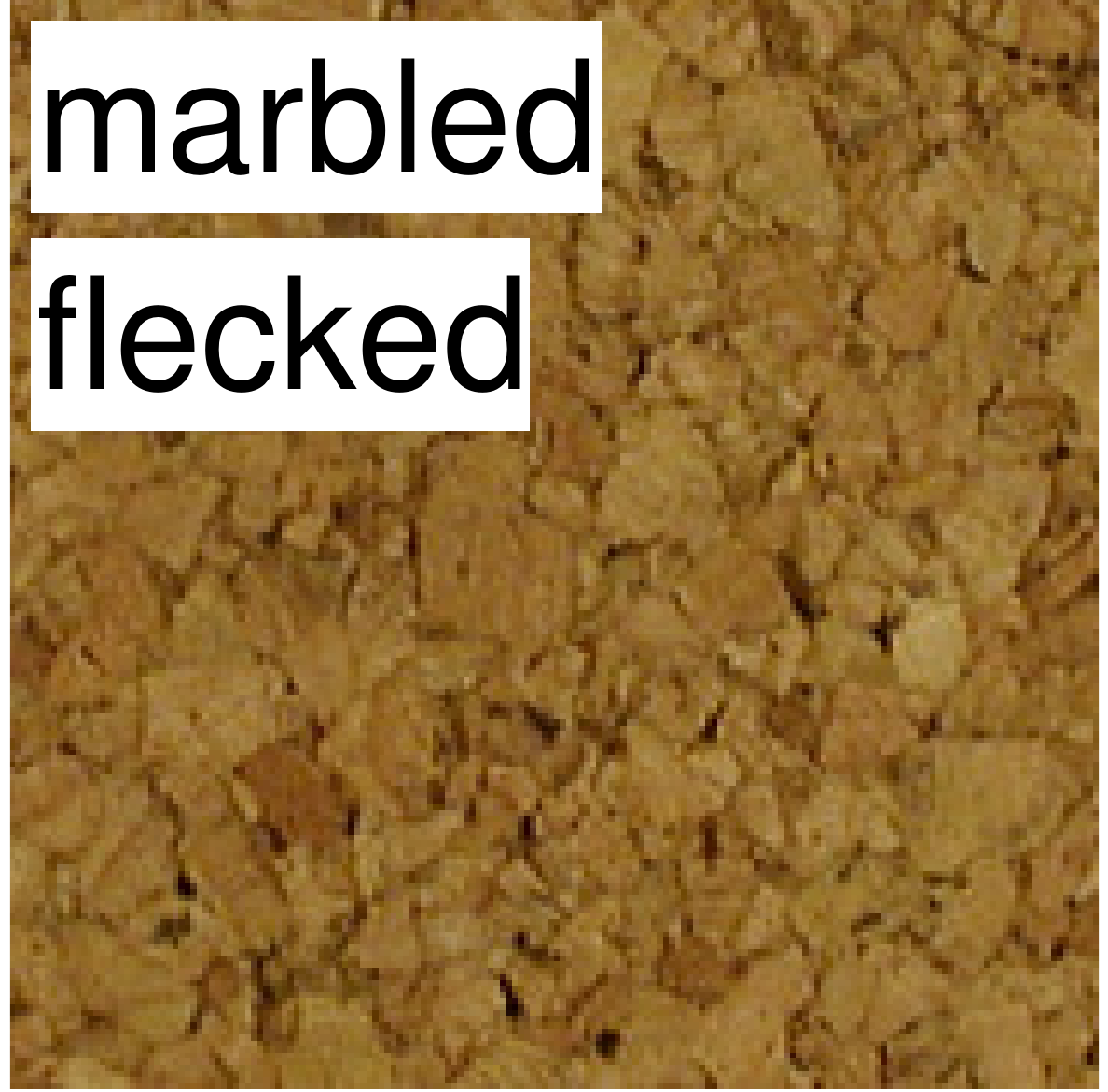}}
	\hfill
	\subfloat[cracker]{\includegraphics[width=\dim]{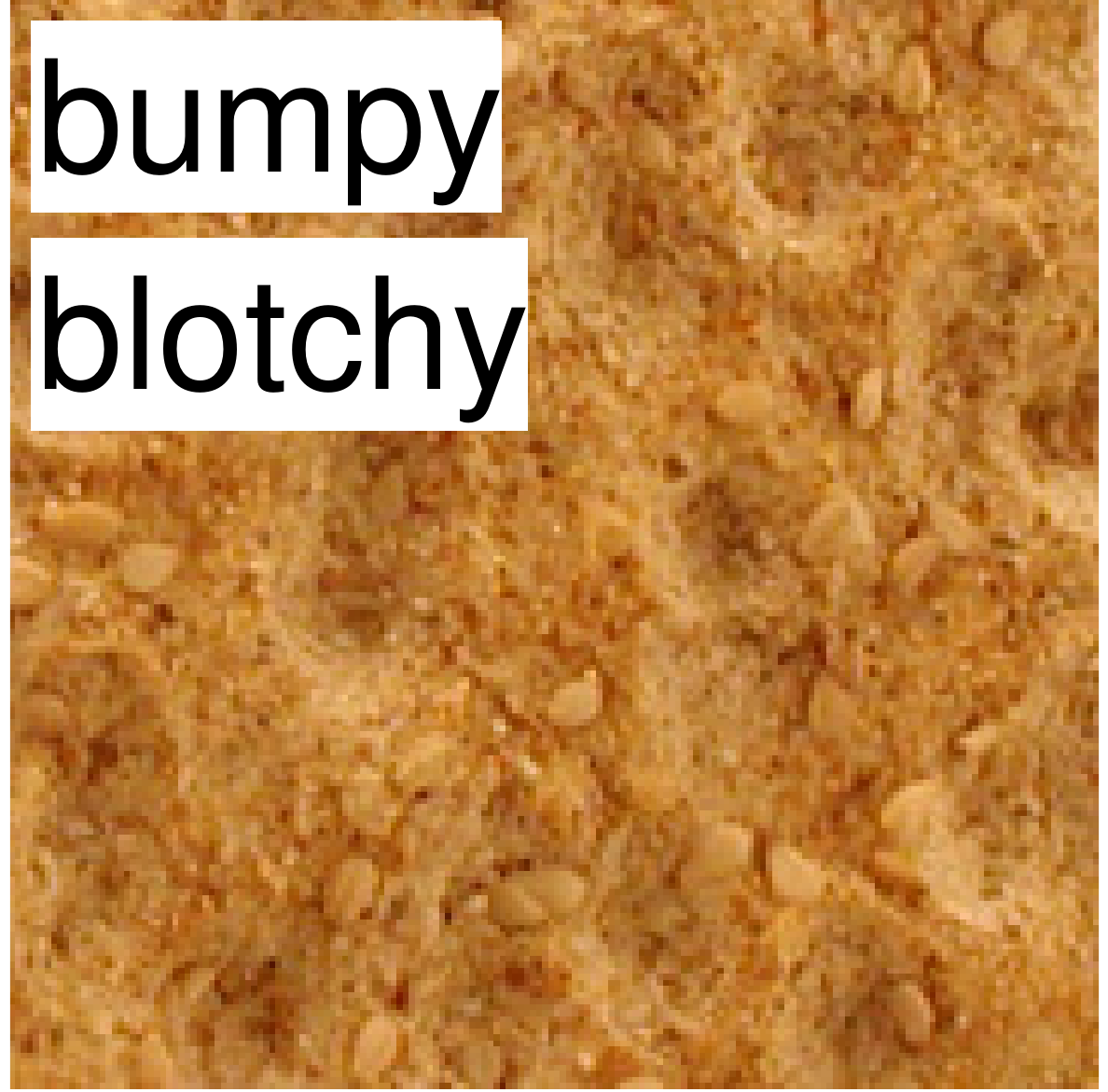}}
	\hfill
	\subfloat[lettuce leaf]{\includegraphics[width=\dim]{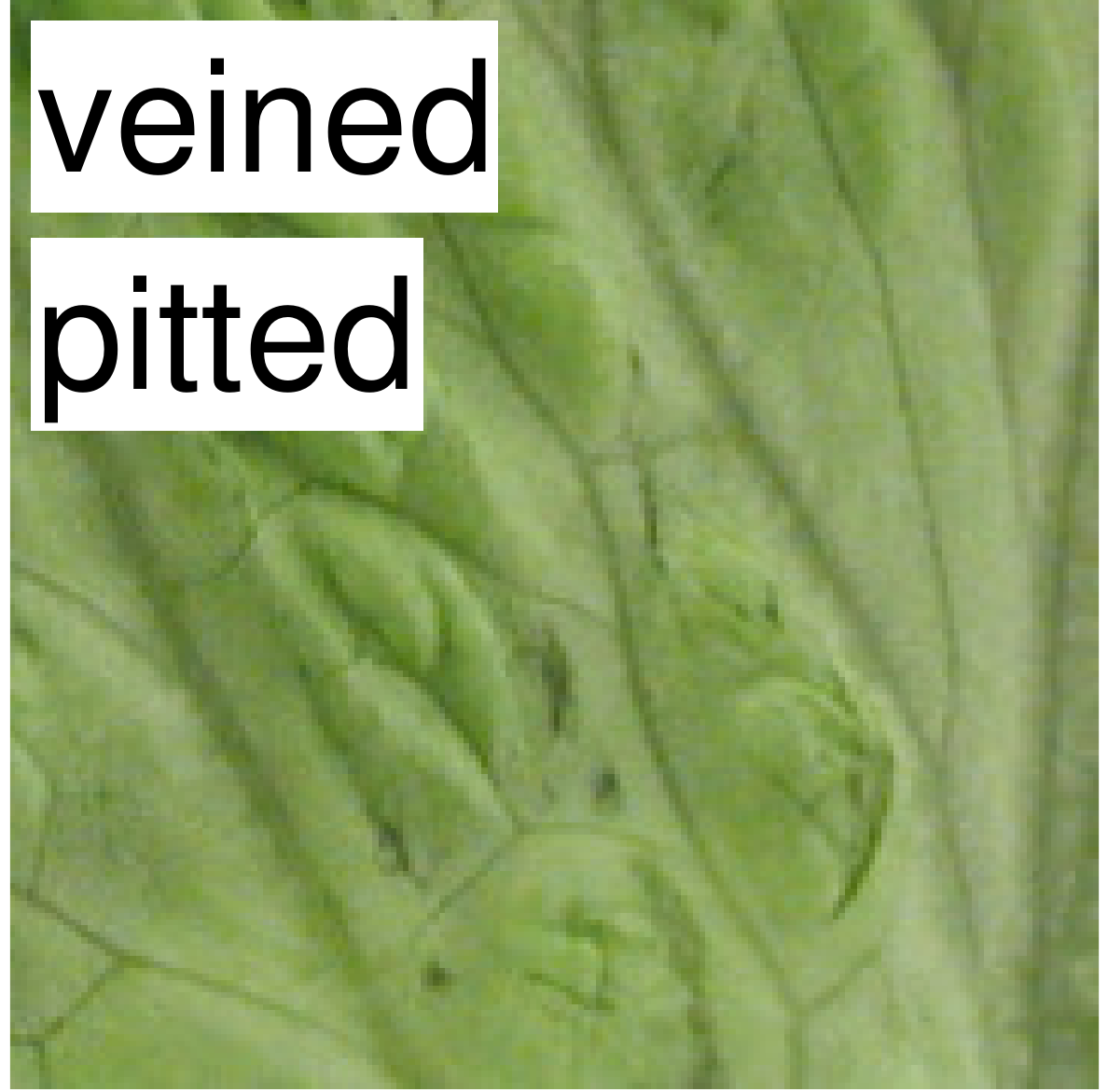}}
	\hfill
	\subfloat[linen]{\includegraphics[width=\dim]{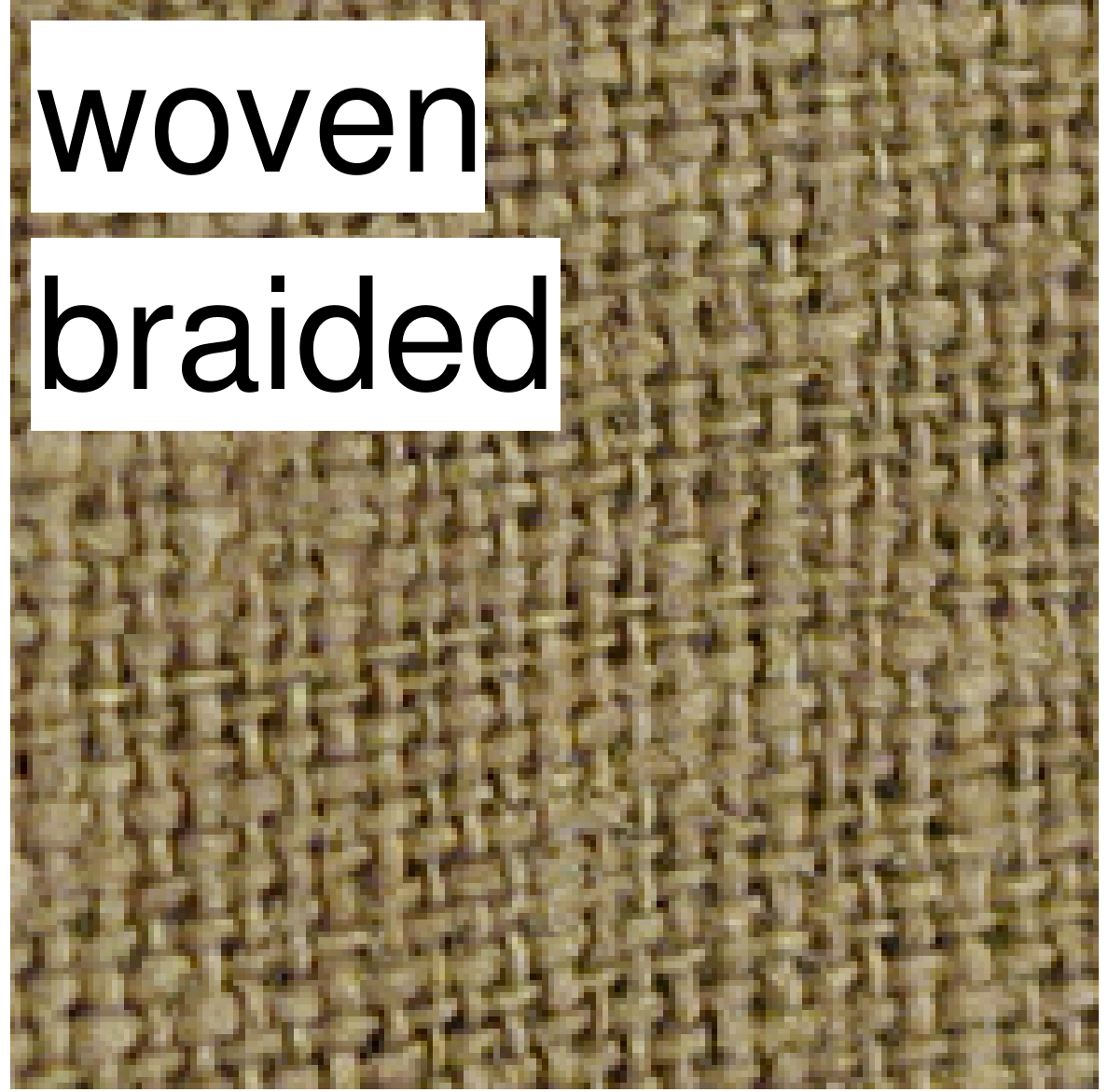}}
	\hfill
	\subfloat[white bread]{\includegraphics[width=\dim]{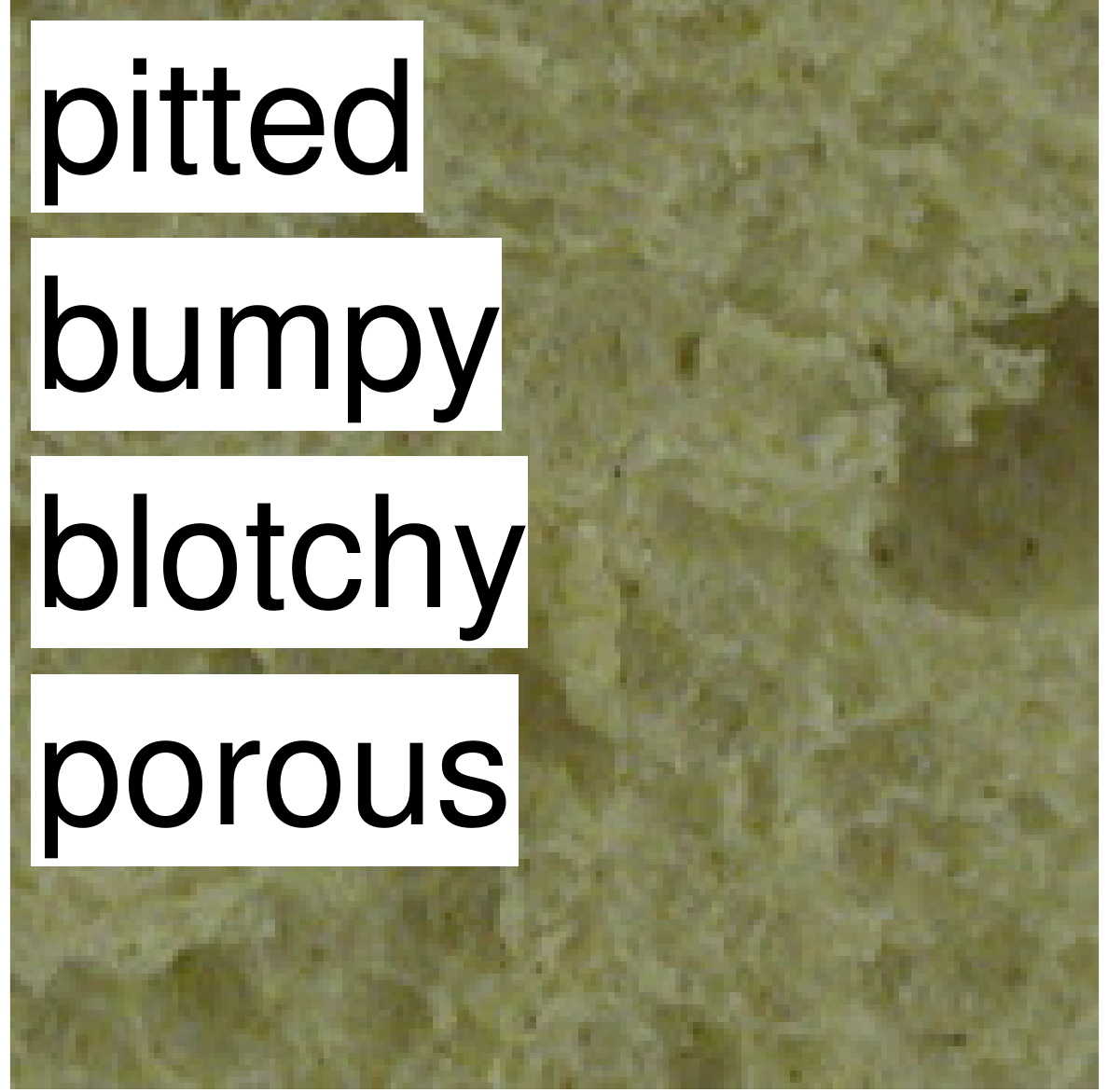}}
	\hfill
	\subfloat[wood]{\includegraphics[width=\dim]{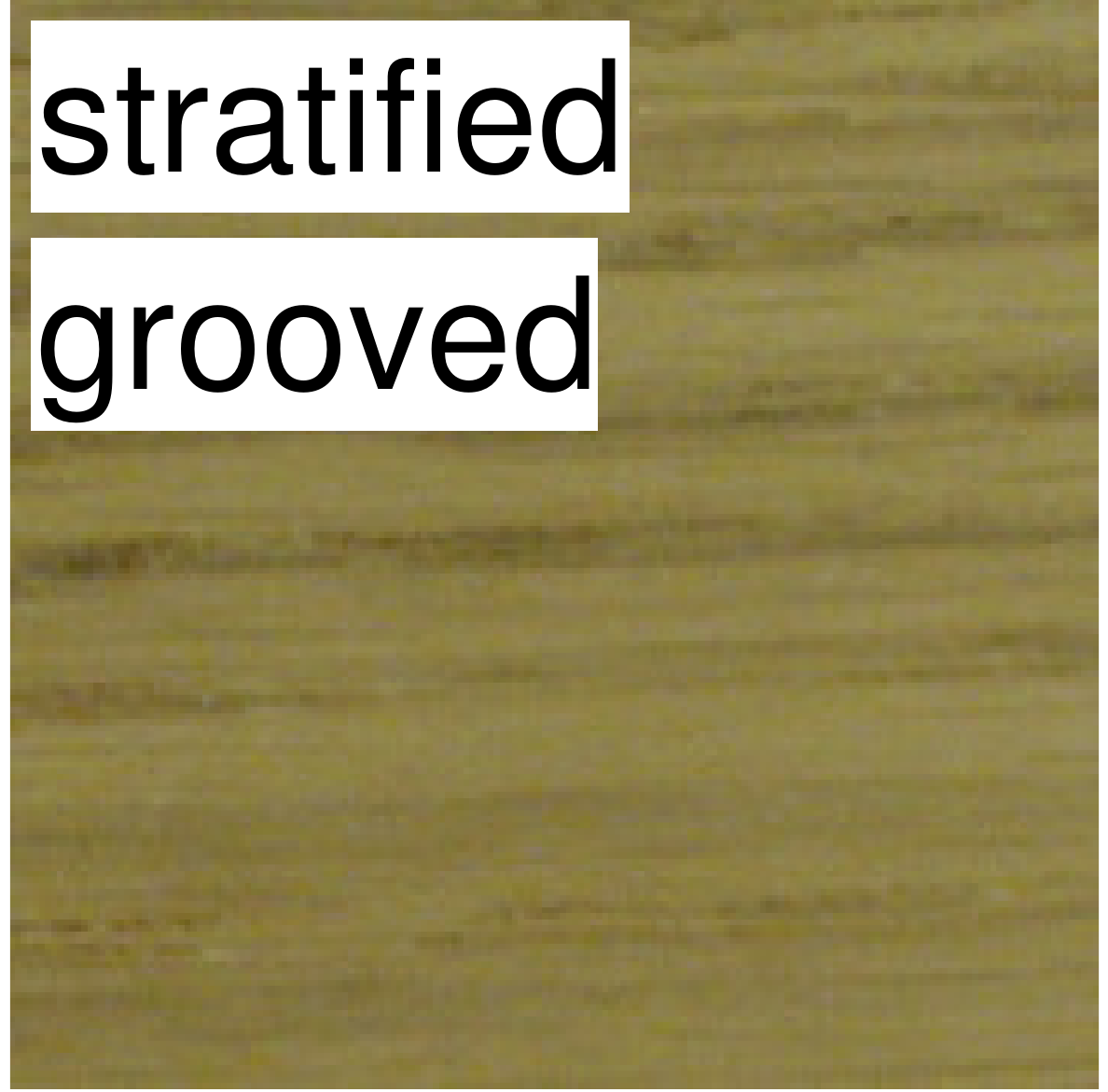}}
	\hfill
	\subfloat[wool]{\includegraphics[width=\dim]{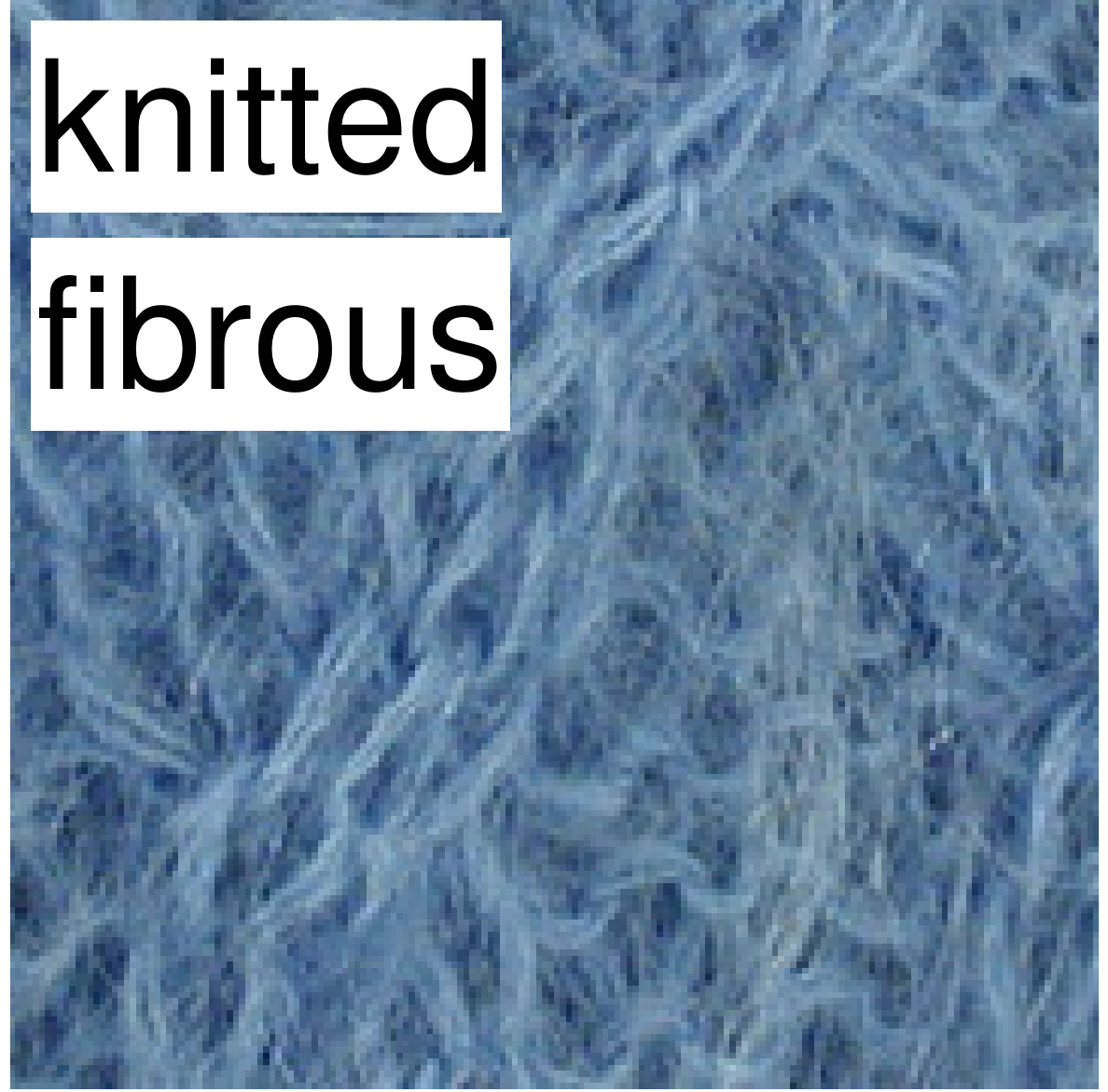}}
\caption{Descriptions of materials from KTH-TIPS-2b dataset. These words are the most frequent top scoring texture attributes (from the list of 47 we proposed), when classifying the images from the KTH-TIPS-2b dataset.}
\label{f:kth-tips-described}
\end{figure*}

\subsection{Search and visualization}\label{s:exp3}

Fig.~\ref{f:kth-tips-described} shows that there is an excellent semantic correlation between the ten categories in KTH-TIPS-2b and the attributes in DTD. For example, aluminium foil is found to be \emph{wrinkled}, while bread is found as:  \emph{bumpy}, \emph{pitted}, \emph{porous} and \emph{flecked}.

In what follows, we experimented with describing images from a challenging material dataset, FMD and encouraged by the good results, we applied the same technique to images from the wild, from some online catalog.

\subsubsection{Subcategorizing FMD materials using describable texture attributes}\label{s:fmd}

\begin{figure}[hb!]
\centering
\includegraphics[width=0.49\textwidth]{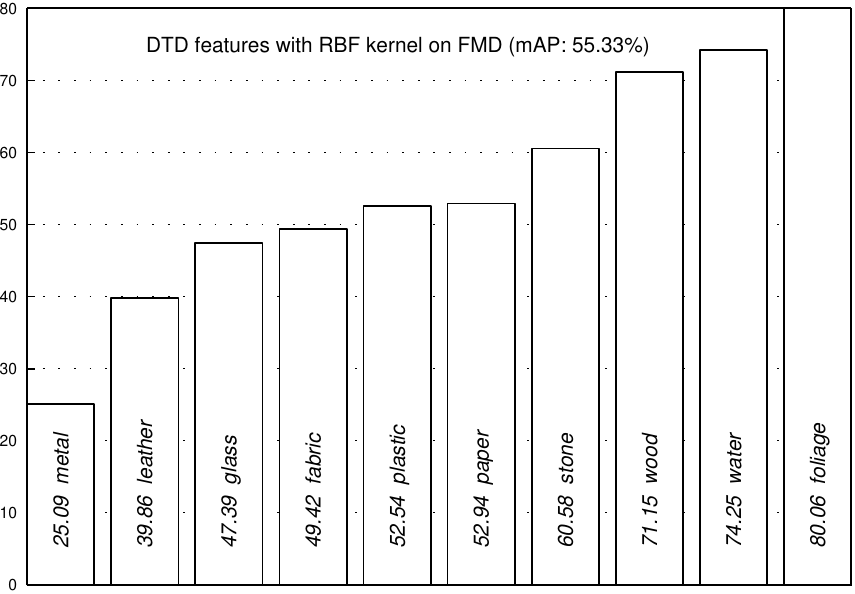}
\caption{Per class AP results on FMD dataset using DTD classification scores as features.
\label{fig:fmd-ap}}
\end{figure}

The results shown in Fig.~\ref{fig:fmd-ap} extends the results in Table~\ref{tbl:dataset-results} and Sect.~\ref{s:exp2} and illustrate the classification performance of the 47-dimensional DTD descriptors on the FMD materials -- note the excellent performance obtained for foliage, wood, and water, which are above 70\%.

Our experiments illustrate how the DTD attributes can be used to find ``semantic structures'' in a dataset such as FMD, for example by distinguishing between ``knitted vs pleated fabric'', ``gauzy vs crystalline glass'', ``veined vs frilly foliage'' etc.  To do so, FDM images for each material were clustered based on the 47 attribute vectors using $K$-means into 3-5 clusters each. Examples of the most meaningful clusters are shown in Fig.~\ref{fig:fmd-clusters} along with the dominant attributes in each.

Notable fine-grained material distinctions include \emph{knitted} vs \emph{pleated} fabrics and \emph{frilly} vs \emph{pleated \& veined} foliage which contain linear structures. In the latter case, veins often have a radial pattern which is captured by the dominant \emph{spiralled} attribute. The method distinguishes \emph{bumpy} stones such as pebbles from \emph{porous} or \emph{pitted} stones for zoomed / detailed views of stone blocks. Water is divided into \emph{swirly \& spiralled} images, which show the orientation of the waves, and \emph{bubbly, sprinkled} images, which show splashing drops. Glass is more challenging but some images are correctly identified as \emph{crystalline}. Fig.~\ref{fig:challenging} shows other challenging examples illustrating the variety of materials and patterns that can be described by the DTD attributes. Metal is one of the hardest class to identify (Fig.~\ref{fig:fmd-ap}), but attributes such as ``interlaced'' and ``braided'' are still correctly recognized in the third (jewelry) and last (metal wires) image.

{
\renewcommand{\dim}{0.144\textwidth}
\begin{figure*}[!t]
\centering
\includegraphics[width=\dim, height=\dim]{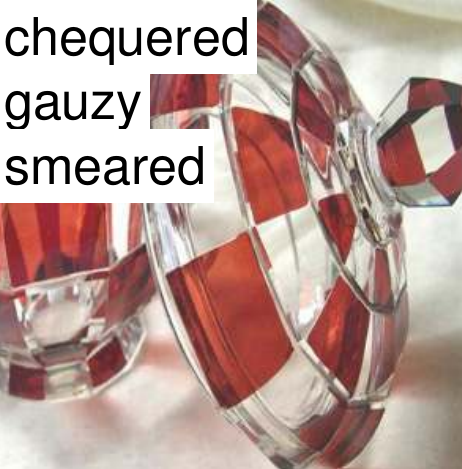}
~
\includegraphics[width=\dim, height=\dim]{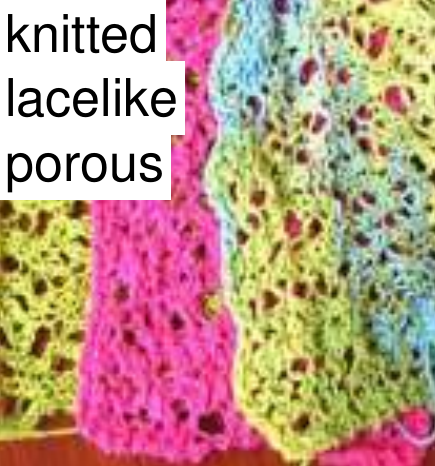} 
~
\includegraphics[width=\dim, height=\dim]{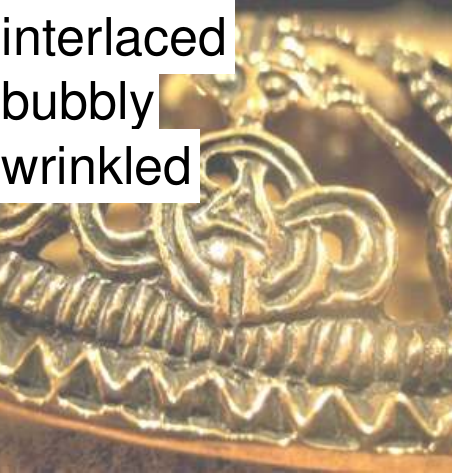}
~
\includegraphics[width=\dim, height=\dim]{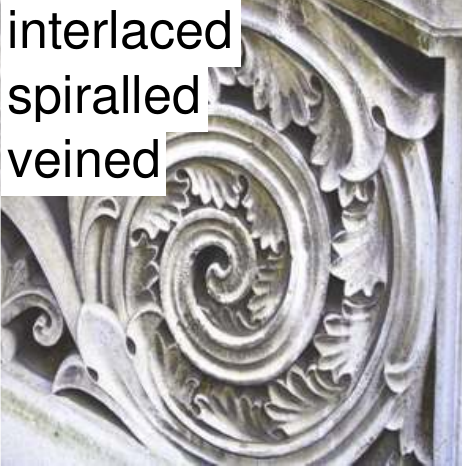}
~
\includegraphics[width=\dim, height=\dim]{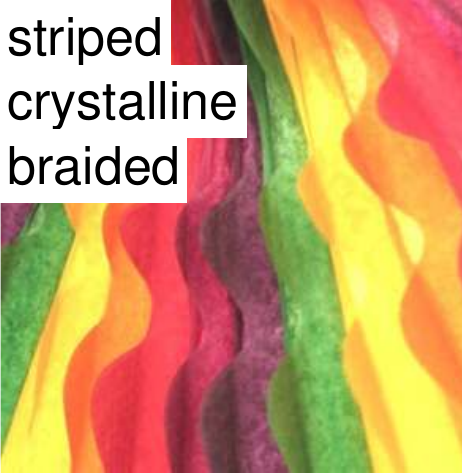}
~
\includegraphics[width=\dim]{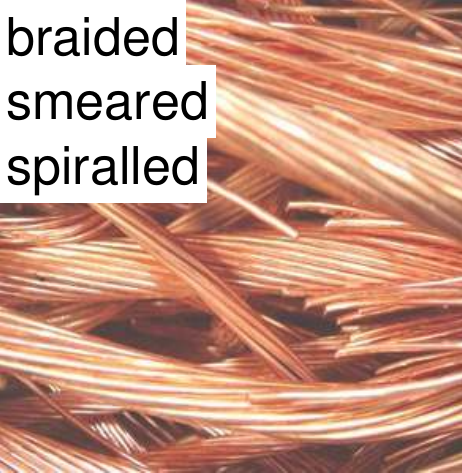}
\caption{Challenging or difficult images which were correctly characterized by our DTD classifier.
\label{fig:challenging}}
\end{figure*}
}

\subsubsection{Examples in the wild}
\begin{figure*}[t]
\includegraphics[width=\textwidth]{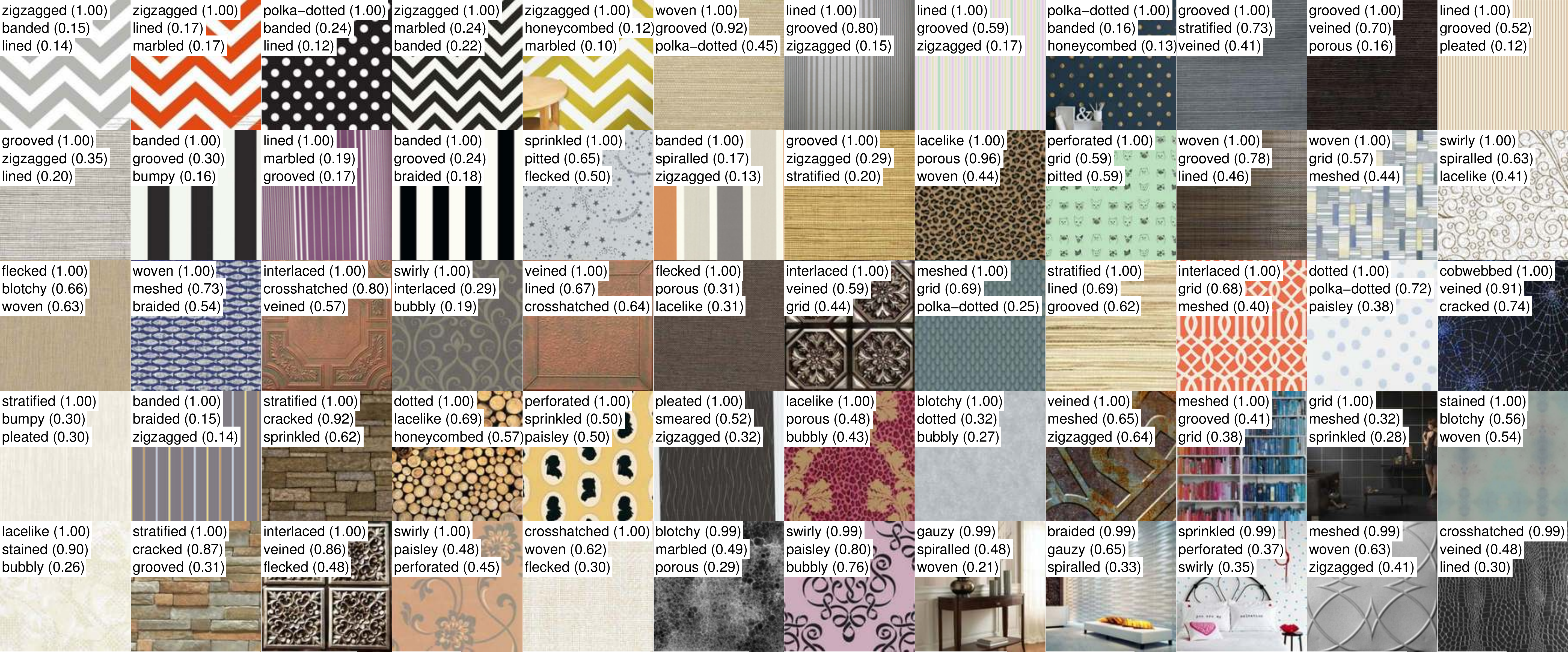}
\caption{Example wallpaper images from an online catalog (houzz.com).
\label{fig:wallpapers}}
\end{figure*}

\begin{figure*}[t]
\includegraphics[width=\textwidth]{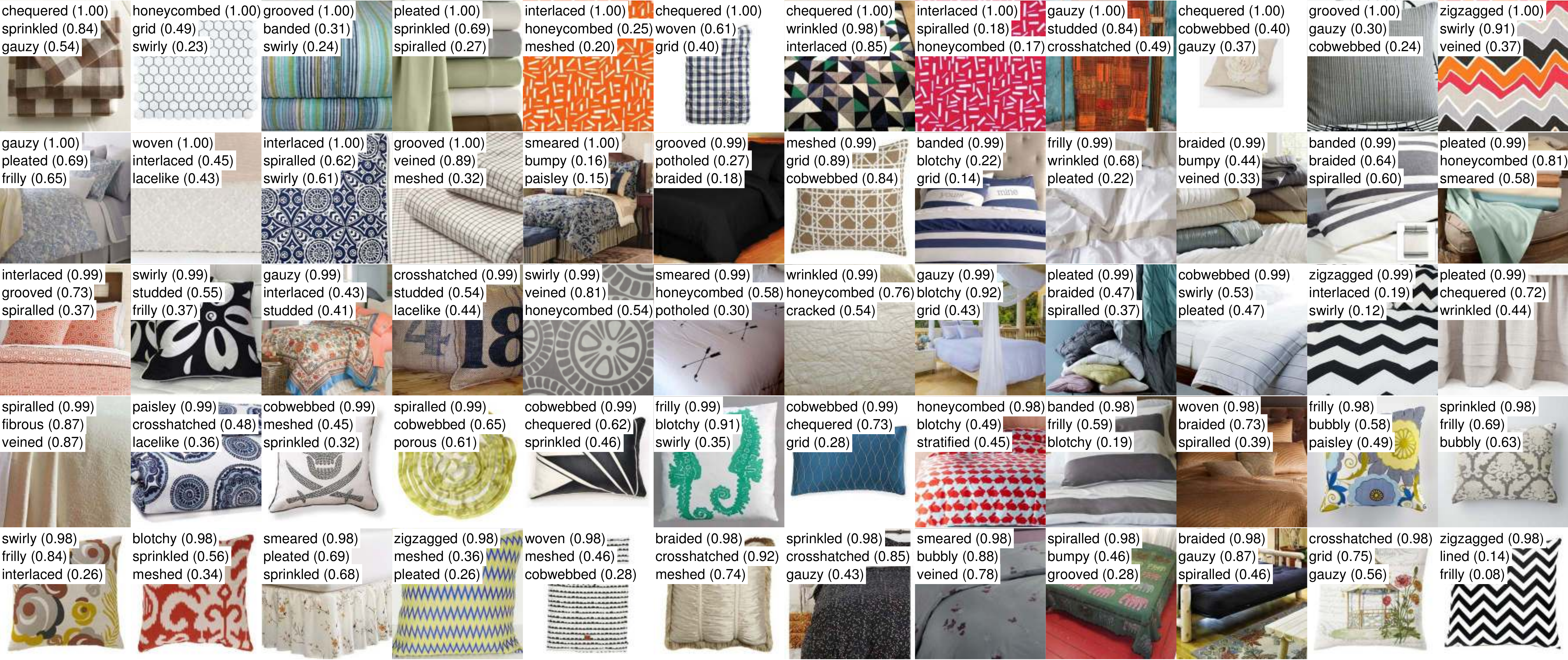}
\caption{Example bedding sets from an online catalog (houzz.com).
\label{fig:beddings}}
\end{figure*}

As an additional application of our describable texture attributes we compute them on a large dataset of 10,000 wallpapers and bedding sets (about 5,000 for each of the two categories) from \url{houzz.com}. The 47 attribute classifiers are learned as explained in Sect.~\ref{s:exp1} using the \IFVSIFT representation and them apply them to the 10,000 images to predict the strength of association of each attribute and image. Classifiers scores are recalibrated on a subset of the target data and converted to probabilities using Platt's method~\cite{platt00probabilistic}, for each individual attribute. Fig.~\ref{fig:wallpapers} and Fig.~\ref{fig:beddings} shows some example attribute predictions (excluding images used for calibrating the scores), for the best scoring 3-4 images for each category -- by top attribute. We show for each images the top three attributes -- the top two being very accurate, while the third is correct in about half of the cases. Please note that each score is calibrated on a per attribute basis, to the scores do not add up to 1.

\section{Summary}

We introduced a large dataset of 5,640 images collected ``in the wild'' jointly labeled with 47 describable texture attributes and used it to study the problem of extracting semantic properties of textures and patterns, addressing real-world human-centric applications. Looking for the best representation to recognize such describable attributes in natural images, we have ported IFV, an object recognition representation, to the texture domain. Not only IFV works best in recognizing describable attributes, but it also outperforms specialized texture representation on a number of challenging material recognition benchmarks. We have shown that the describable attributes, while not being designed to do so, are good predictors of materials as well, and that, when combined with IFV, significantly outperform the state-of-the-art on the FMD and KTH-TIPS recognition tasks.

\footnotesize
\bibliographystyle{ieee}
\bibliography{references}

\begin{thebibliography}{10}\itemsep=-1pt

\bibitem{amadasun89textural}
M.~Amadasun and R.~King.
\newblock Textural features corresponding to textural properties.
\newblock {\em Systems, Man, and Cybernetics}, 19(5), 1989.

\bibitem{bajcsy73computer}
R.~Bajcsy.
\newblock Computer description of textured surfaces.
\newblock In {\em IJCAI}, IJCAI. Morgan Kaufmann Publishers Inc., 1973.

\bibitem{berg2010automatic}
T.~Berg, A.~Berg, and J.~Shih.
\newblock Automatic attribute discovery and characterization from noisy web
  data.
\newblock {\em ECCV}, 2010.

\bibitem{berlin1991basic}
B.~Berlin and P.~Kay.
\newblock {\em Basic color terms: Their universality and evolution}.
\newblock Univ of California Press, 1991.

\bibitem{bhushan1997texture}
N.~Bhushan, A.~Rao, and G.~Lohse.
\newblock The texture lexicon: Understanding the categorization of visual
  texture terms and their relationship to texture images.
\newblock {\em Cognitive Science}, 21(2):219--246, 1997.

\bibitem{bourdev2011describing}
L.~Bourdev, S.~Maji, and J.~Malik.
\newblock Describing people: A poselet-based approach to attribute
  classification.
\newblock In {\em ICCV}, 2011.

\bibitem{caputo05class}
B.~Caputo, E.~Hayman, and P.~Mallikarjuna.
\newblock Class-specific material categorisation.
\newblock In {\em ICCV}, 2005.

\bibitem{csurka04visual}
G.~Csurka, C.~R. Dance, L.~Dan, J.~Willamowski, and C.~Bray.
\newblock Visual categorization with bags of keypoints.
\newblock In {\em Proc. {ECCV} Workshop on Stat. Learn. in Comp. Vision}, 2004.

\bibitem{dana99reflectance}
K.~J. Dana, B.~van Ginneken, S.~K. Nayar, and J.~J. Koenderink.
\newblock Reflectance and texture of real world surfaces.
\newblock {\em ACM Transactions on Graphics}, 18(1):1--34, 1999.

\bibitem{deng09imagenet}
J.~Deng, W.~Dong, R.~Socher, L.-J. Li, K.~Li, and L.~Fei-Fei.
\newblock {ImageNet: A Large-Scale Hierarchical Image Database}.
\newblock In {\em CVPR}, 2009.

\bibitem{farhadi2009describing}
A.~Farhadi, I.~Endres, D.~Hoiem, and D.~Forsyth.
\newblock Describing objects by their attributes.
\newblock In {\em CVPR}, pages 1778--1785. IEEE, 2009.

\bibitem{ferrari2008learning}
V.~Ferrari and A.~Zisserman.
\newblock Learning visual attributes.
\newblock In {\em NIPS}, 2007.

\bibitem{geusebroek03fast}
J.~M. Geusebroek, A.~W.~M. Smeulders, and J.~van~de Weijer.
\newblock Fast anisotropic gauss filtering.
\newblock {\em IEEE Transactions on Image Processing}, 12(8):938--943, 2003.

\bibitem{hayman04learning}
E.~Hayman, B.~Caputo, M.~Fritz, and J.-O. Eklundh.
\newblock On the significance of real-world conditions for material
  classification.
\newblock {\em ECCV}, 2004.

\bibitem{jegou10aggregating}
H.~J{\'e}gou, M.~Douze, C.~Schmid, and P.~P{\'e}rez.
\newblock Aggregating local descriptors into a compact image representation.
\newblock In {\em Proc. {CVPR}}, 2010.

\bibitem{julesz81textons}
B.~Julesz.
\newblock Textons, the elements of texture perception, and their interactions.
\newblock {\em Nature}, 290(5802):91--97, march 1981.

\bibitem{kumar2011describable}
N.~Kumar, A.~Berg, P.~Belhumeur, and S.~Nayar.
\newblock Describable visual attributes for face verification and image search.
\newblock {\em PAMI}, 33(10):1962--1977, 2011.

\bibitem{lazebnik05sparse}
S.~Lazebnik, C.~Schmid, and J.~Ponce.
\newblock A sparse texture representation using local affine regions.
\newblock {\em PAMI}, 28(8):2169--2178, 2005.

\bibitem{leung2001representing}
T.~Leung and J.~Malik.
\newblock Representing and recognizing the visual appearance of materials using
  three-dimensional textons.
\newblock {\em International Journal of Computer Vision}, 43(1):29--44, 2001.

\bibitem{lowe99object}
D.~G. Lowe.
\newblock Object recognition from local scale-invariant features.
\newblock In {\em Proc. {ICCV}}, 1999.

\bibitem{malik90preattentive}
J.~Malik and P.~Perona.
\newblock Preattentive texture discrimination with early vision mechanisms.
\newblock {\em {JOSA A}}, 7(5), 1990.

\bibitem{matthews13enriching}
T.~Matthews, M.~S. Nixon, and M.~Niranjan.
\newblock Enriching texture analysis with semantic data.
\newblock In {\em CVPR}, June 2013.

\bibitem{ojala2002multiresolution}
T.~Ojala, M.~Pietikainen, and T.~Maenpaa.
\newblock Multiresolution gray-scale and rotation invariant texture
  classification with local binary patterns.
\newblock {\em PAMI}, 24(7):971--987, 2002.

\bibitem{oxholm2012texture}
G.~Oxholm, P.~Bariya, and K.~Nishino.
\newblock The scale of geometric texture.
\newblock In {\em European Conference on Computer Vision}, pages 58--71.
  Springer Berlin/Heidelberg, 2012.

\bibitem{patterson2012sun}
G.~Patterson and J.~Hays.
\newblock Sun attribute database: Discovering, annotating, and recognizing
  scene attributes.
\newblock In {\em CVPR}, 2012.

\break

\bibitem{perronnin07fisher}
F.~Perronnin and C.~R. Dance.
\newblock Fisher kernels on visual vocabularies for image categorization.
\newblock In {\em CVPR}, 2007.

\bibitem{perronnin10improving}
F.~Perronnin, J.~S{\'a}nchez, and T.~Mensink.
\newblock Improving the {F}isher kernel for large-scale image classification.
\newblock In {\em Proc. {ECCV}}, 2010.

\bibitem{platt00probabilistic}
J.~C. Platt.
\newblock Probabilistic outputs for support vector machines and comparisons to
  regularized likelihood methods.
\newblock In A.~Smola, P.~Bartlett, B.~Sch{\"o}lkopf, and D.~Schuurmans,
  editors, {\em Advances in Large Margin Classifiers}. Cambridge, 2000.

\bibitem{sharan13recognizing}
L.~Sharan, C.~Liu, R.~Rosenholtz, and E.~H. Adelson.
\newblock Recognizing materials using perceptually inspired features.
\newblock {\em International Journal of Computer Vision}, 103(3):348--371,
  2013.

\bibitem{sharan09material}
L.~Sharan, R.~Rosenholtz, and E.~H. Adelson.
\newblock Material perceprion: What can you see in a brief glance?
\newblock {\em Journal of Vision}, 9:784(8), 2009.

\bibitem{sharma12local}
G.~Sharma, S.~ul~Hussain, and F.~Jurie.
\newblock Local higher-order statistics (lhs) for texture categorization and
  facial analysis.
\newblock In {\em Proc. {ECCV}}. 2012.

\bibitem{sifre13rotation}
L.~Sifre and S.~Mallat.
\newblock Rotation, scaling and deformation invariant scattering for texture
  discrimination.
\newblock In {\em CVPR}, June 2013.

\bibitem{tamura78textural}
H.~Tamura, S.~Mori, and T.~Yamawaki.
\newblock Textural features corresponding to visual perception.
\newblock {\em Systems, Man and Cybernetics, IEEE Transactions on}, 8(6):460
  --473, june 1978.

\bibitem{timofte12trainingfree}
R.~Timofte and L.~Van~Gool.
\newblock A training-free classification framework for textures, writers, and
  materials.
\newblock In {\em BMVC}, Sept. 2012.

\bibitem{varma2003texture}
M.~Varma and A.~Zisserman.
\newblock Texture classification: Are filter banks necessary?
\newblock In {\em CVPR}, volume~2, pages II--691. IEEE, 2003.

\bibitem{varma2005statistical}
M.~Varma and A.~Zisserman.
\newblock A statistical approach to texture classification from single images.
\newblock {\em IJCV}, 62(1):61--81, 2005.

\bibitem{vedaldi10efficient}
A.~Vedaldi and A.~Zisserman.
\newblock Efficient additive kernels via explicit feature maps.
\newblock In {\em CVPR}, 2010.

\bibitem{welinder10online}
P.~Welinder and P.~Perona.
\newblock Online crowdsourcing: rating annotators and obtaining cost-effective
  labels.
\newblock In {\em CVPR}, 2010.

\bibitem{xu09viewpoint}
Y.~Xu, H.~Ji, and C.~Fermuller.
\newblock Viewpoint invariant texture description using fractal analysis.
\newblock {\em {IJCV}}, 83(1):85--100, jun 2009.

\end{thebibliography}

\renewcommand{\dim}{.97\textwidth}
\newcommand{\nospace}{\hfill\vspace{-0.09in}}

\begin{figure*}[t]
\vspace{-0.1in}
\centering
\subfloat[fabric (knitted)]{\includegraphics[width=\dim]{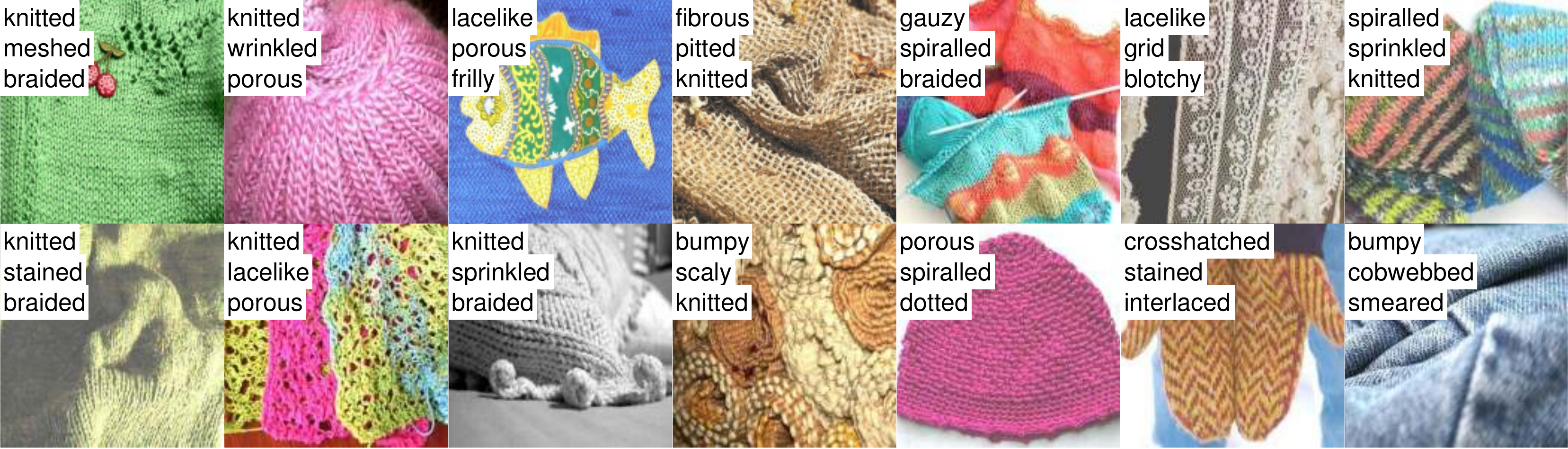}}
\nospace
\subfloat[fabric (pleated)]{\includegraphics[width=\dim]{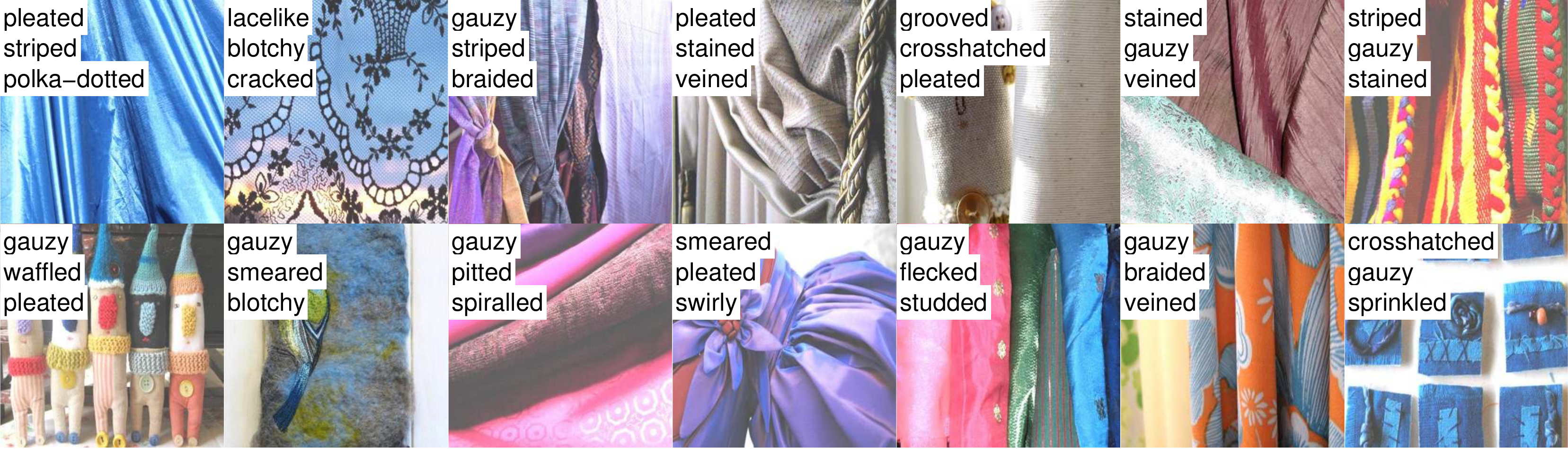}}
\hfill
\nospace
\subfloat[glass (bubbly, gauzy)]{\includegraphics[width=0.55\textwidth]{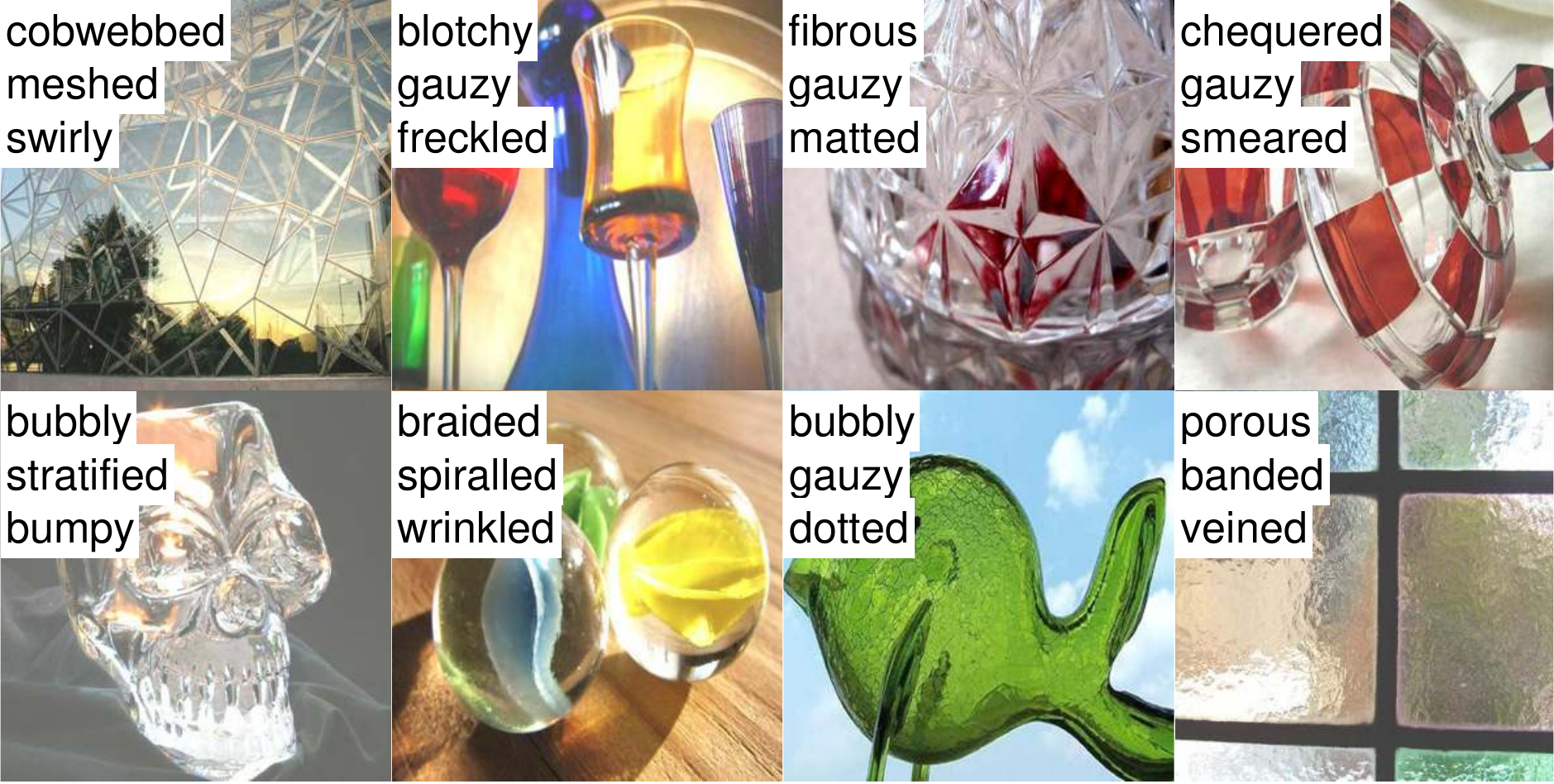}} \hspace{0.1em}
\subfloat[glass (crystalline, bubbly)]{\includegraphics[width=0.42\textwidth]{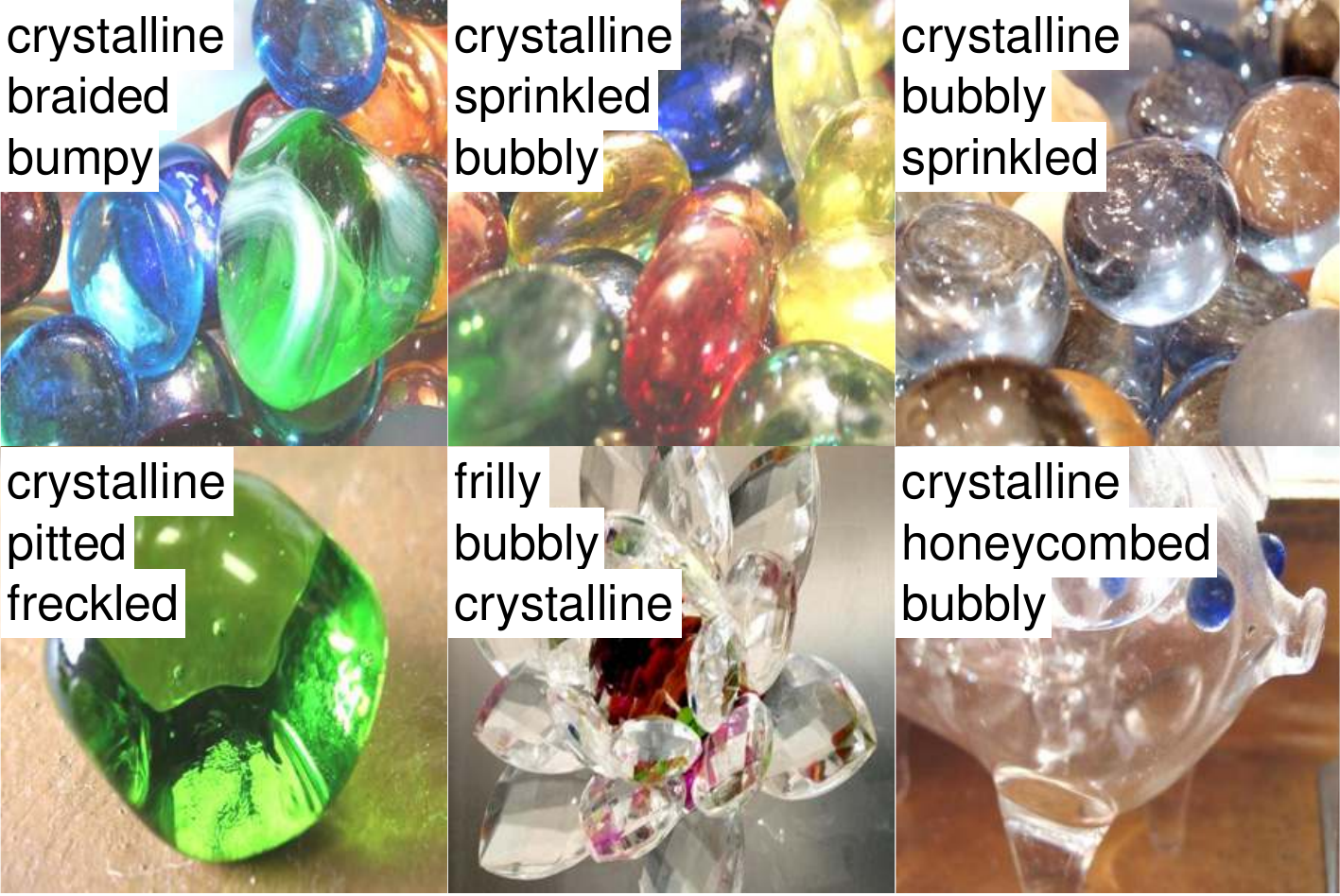}}
\hfill
\caption{Example meaningful clusters of FMD categories, obtained using K-means on DTD classification scores. Showing results for fabric and glass -- overlayed, we list the most frequently identified attributes. On each image, we show the top 3 scoring texture words.} 
\label{fig:fmd-clusters}
\end{figure*}

\begin{figure*}[t]
\centering
\subfloat[foliage (pleated, spiralled, veined)]{\includegraphics[width=\dim]{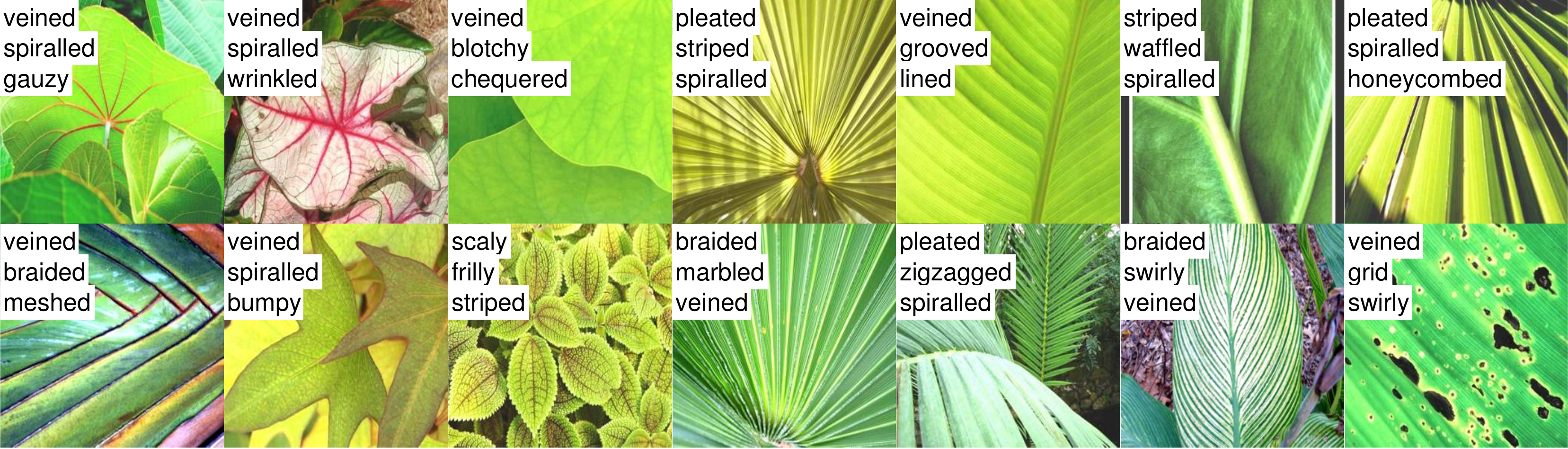}}
\hfill
\subfloat[foliage (frilly, sprinkled)]{\includegraphics[width=\dim]{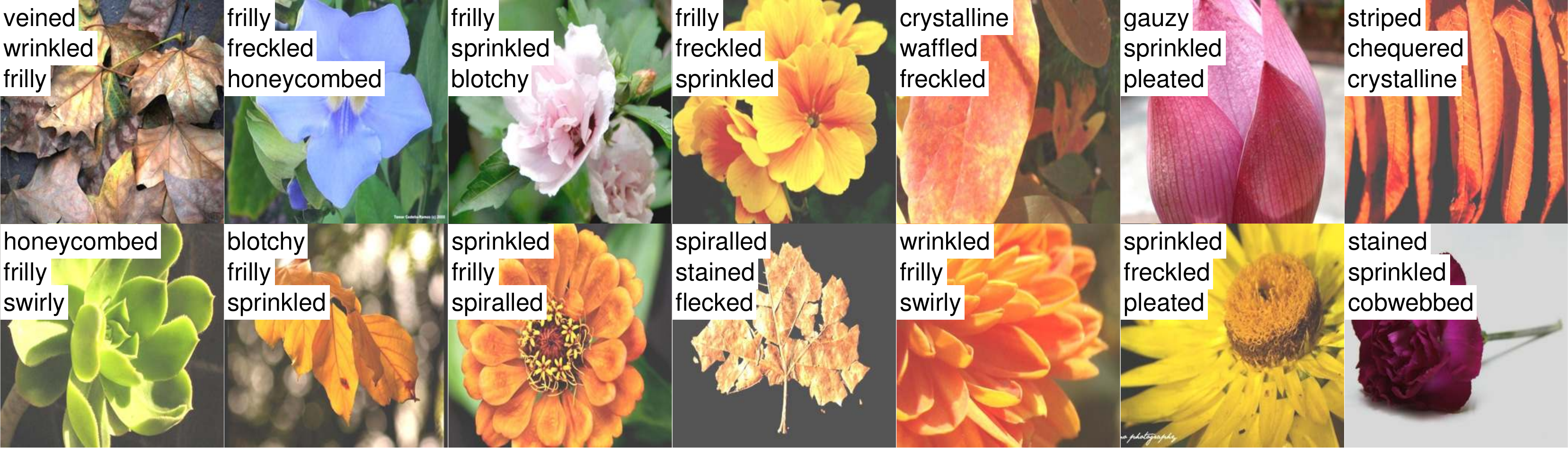}}
\hfill
\subfloat[paper (wrinkled, pleated)]{\includegraphics[width=\dim]{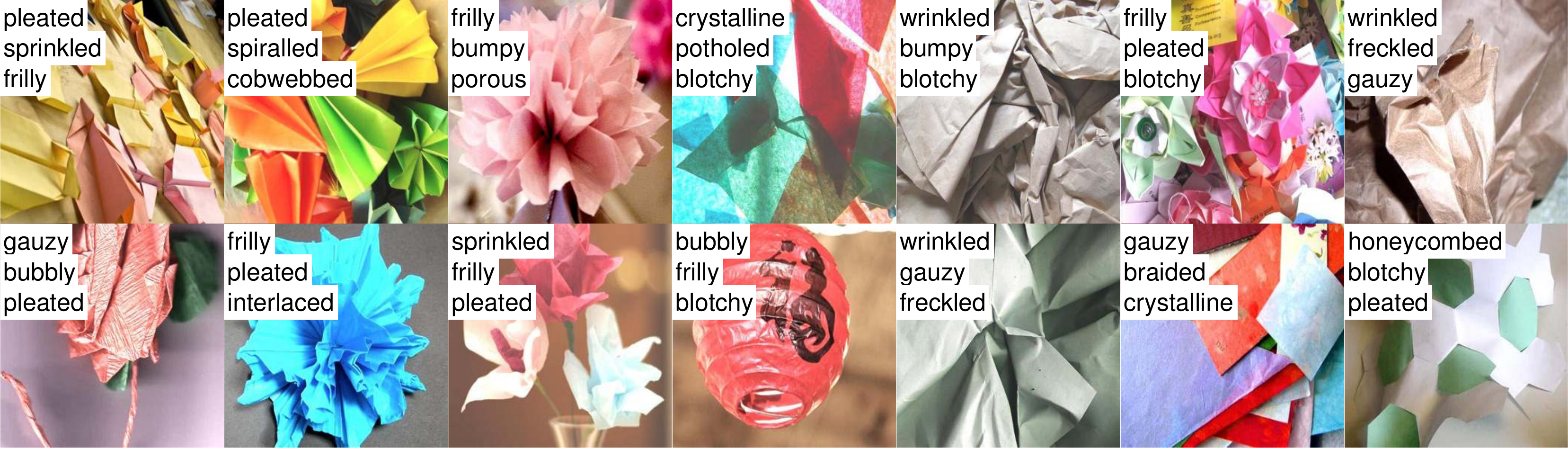}}
\hfill
\subfloat[wood (cracked, veined, interlaced)]{\includegraphics[width=\dim]{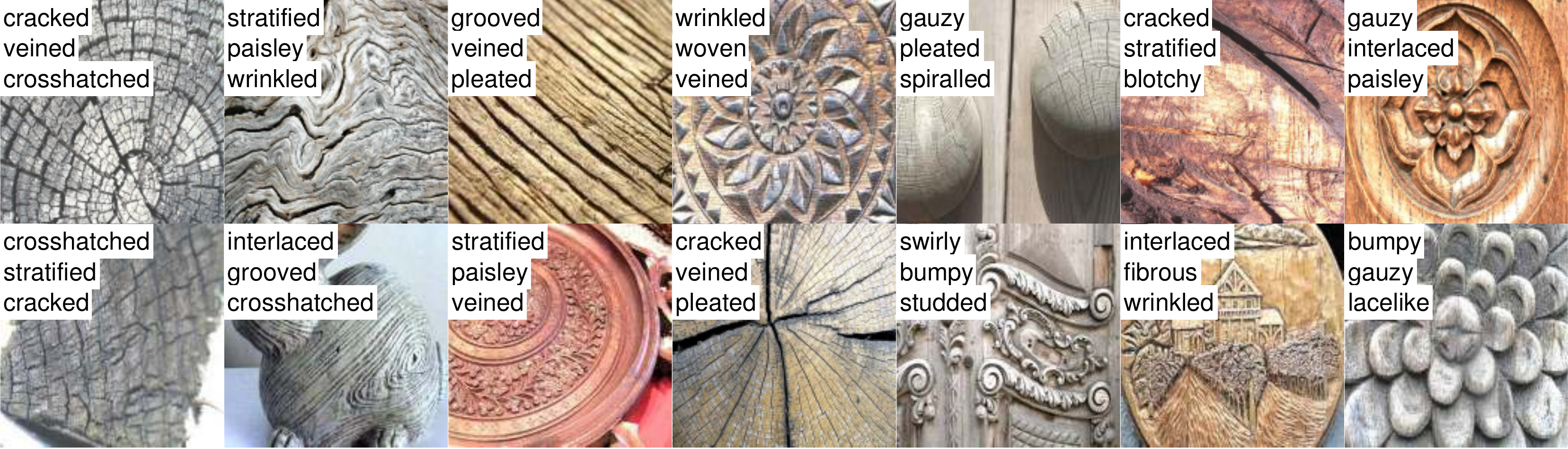}}
\caption[]{\emph{Continued from Fig.~\ref{fig:fmd-clusters}}. Displaying results on foliage, paper and wood.
\label{fig:fmd-clusters-2}}
\end{figure*}

\begin{figure*}[t]
\subfloat[stone (bumpy)]{\includegraphics[width=\dim]{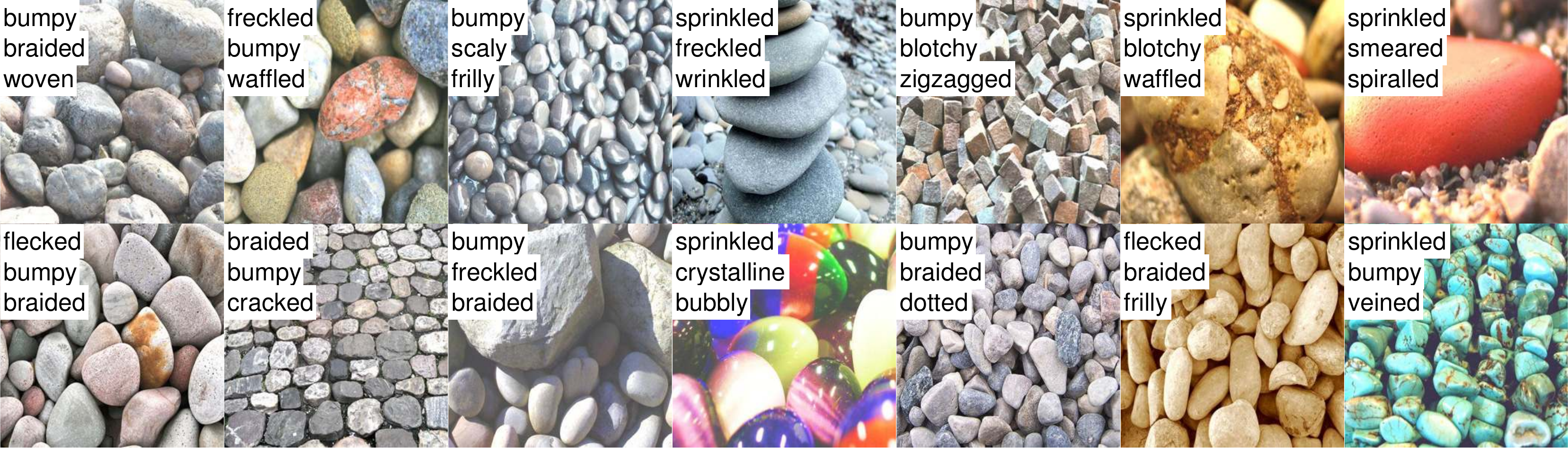}}
\hfill
\vspace{-0.05in}
\subfloat[stone (porous, pitted, flecked)]{\includegraphics[width=\dim]{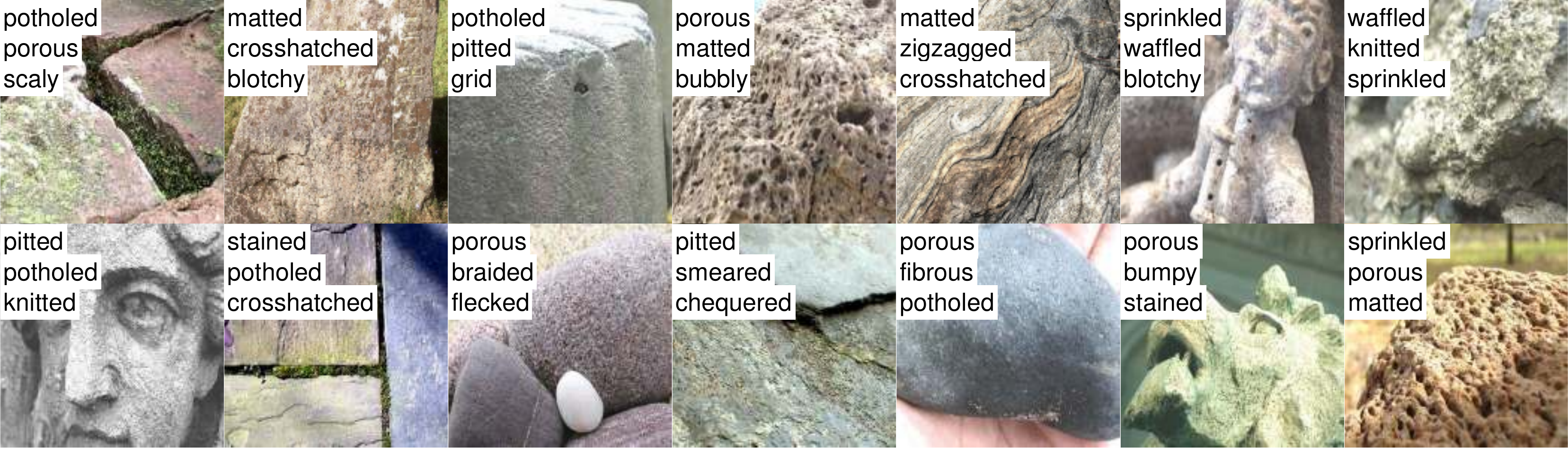}}
\nospace
\subfloat[water (bubbly, smeared)]{\includegraphics[width=\dim]{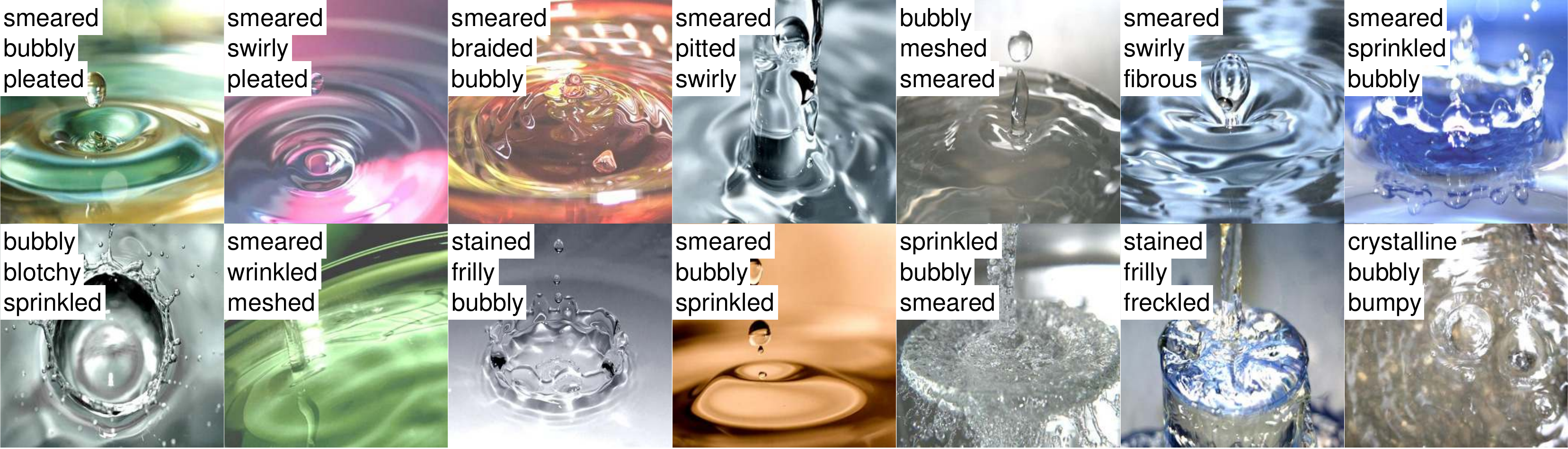}}
\nospace
\subfloat[water (swirly, spiralled)]{\includegraphics[width=\dim]{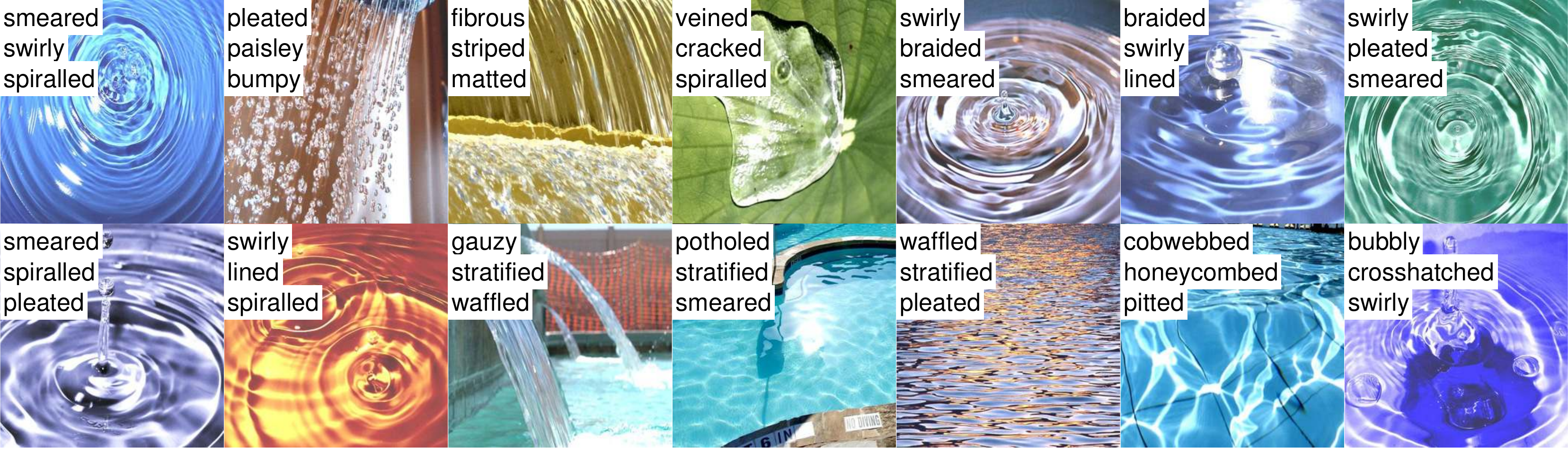}}
\caption[]{ \emph{Continued from Fig.~\ref{fig:fmd-clusters-2}} Subcategories for stone and water images.
\label{fig:fmd-clusters-3}}
\end{figure*}

\end{document}